\documentclass[10pt,twocolumn,letterpaper]{article}

\usepackage{iccv}
\usepackage{times}
\usepackage{epsfig}
\usepackage{graphicx}
\usepackage{amsmath}
\usepackage{amssymb}

\usepackage{array}
\usepackage[table]{xcolor}
\usepackage{colortbl}
\usepackage{pifont}
\usepackage{cite}
\usepackage{diagbox}
\usepackage{rotating}
\usepackage{booktabs}
\usepackage{overpic}
\usepackage{textcomp}
\usepackage{contour}

\usepackage[accsupp]{axessibility}  

\def\ie{\emph{i.e.}}
\def\eg{\emph{e.g.}}

\def\etal{{\em et al.~}}
\def\ours{\emph{SP-Net}}

\newlength\savedwidth
\newcommand{\whline}[1]{\noalign{\global\savedwidth\arrayrulewidth \global\arrayrulewidth #1}%
                   \hline \noalign{\global\arrayrulewidth\savedwidth}}

\usepackage[pagebackref=true,breaklinks=true,letterpaper=true,colorlinks,bookmarks=false]{hyperref}

\iccvfinalcopy 

\def\iccvPaperID{****} 

\ificcvfinal\fi

\begin{document}

\title{Specificity-preserving RGB-D Saliency Detection}


\author{Tao Zhou$^{1}$, ~Deng-Ping Fan$^{2}$\thanks{Corresponding author: \emph{Deng-Ping Fan (dengpfan@gmail.com)}}, ~Geng Chen$^{3}$, ~Yi Zhou$^4$, ~Huazhu Fu$^5$\\ 	
	   \normalsize{$^1$School of Computer Science and Engineering, Nanjing University of Science and Technology, China}\\
	   \normalsize{$^2$CVL, ETH Zurich, Switzerland}\\
	   \normalsize{$^3$School of Computer Science and Engineering, Northwestern Polytechnical University, Xi'an, China} \\
	   \normalsize{$^4$School of Computer Science and Engineering, Southeast University, China}\\
	   \normalsize{$^4$Institute of High Performance Computing, Agency for Science, Technology and Research, Singapore}\\
}


\maketitle
\ificcvfinal\thispagestyle{empty}\fi

\begin{abstract}

Salient object detection (SOD) on RGB and depth images has attracted more and more research interests, due to its effectiveness and the fact that depth cues can now be conveniently captured. Existing RGB-D SOD models usually adopt different fusion strategies to learn a shared representation from the two modalities (\ie, RGB and depth), while few methods explicitly consider how to preserve modality-specific characteristics. In this study, we propose a novel framework, termed \emph{\textbf{SPNet}} (\textbf{S}pecificity-\textbf{P}reserving \textbf{N}etwork), which benefits SOD performance by exploring both the shared information and modality-specific properties (\eg, specificity). Specifically, we propose to adopt two modality-specific networks and a shared learning network to generate individual and shared saliency prediction maps, respectively. To effectively fuse cross-modal features in the shared learning network, we propose a cross-enhanced integration module (CIM) and then propagate the fused feature to the next layer for integrating cross-level information. Moreover, to capture rich complementary multi-modal information for boosting the SOD performance, we propose a multi-modal feature aggregation (MFA) module to integrate the modality-specific features from each individual decoder into the shared decoder. By using a skip connection, the hierarchical features between the encoder and decoder layers can be fully combined. Extensive experiments demonstrate that our~\ours~outperforms cutting-edge approaches on six popular RGB-D SOD and three camouflaged object detection benchmarks. The project is publicly available at: \url{https://github.com/taozh2017/SPNet}. 

\end{abstract}


\section{Introduction}

Salient object detection (SOD)\footnote{We use ``saliency detection" \& ``SOD" interchangeably.} aims to model the mechanism of human visual
attention and locate the most visually distinctive object(s) in a given scene~\cite{peng2014rgbd}. SOD has been widely applied in various vision-related tasks, such as image understanding \cite{zhu2014unsupervised}, action recognition \cite{rapantzikos2009dense}, \cite{shimoda2016distinct}, video/semantic segmentation \cite{wang2017saliency,shimoda2016distinct}, and person re-identification \cite{zhao2016person}. Although significant progress has been made, it is still challenging to accurately locate salient objects in many challenging scenarios, such as instance cluttered background, low-contrast lighting conditions, and salient object(s) having a similar appearance with the background. Recently, with the large availability of depth sensors in smart devices, depth maps have been introduced to provide geometric and spatial information to improve SOD performance. Consequently, fusing RGB and depth images has gained increasing interest in the SOD community \cite{fan2019rethinking,cascaded_rgbd_sod,liu2021vst,zhou2021rgb,fu2020jl,zhang2020uc,chen2021cnn,li2021hierarchical,zhao2021rgb}, and it is a challenging task to adaptively fuse the two modalities (\ie, RGB and depth).

\begin{figure}[t]
    \centering
    \includegraphics[width=1\linewidth]{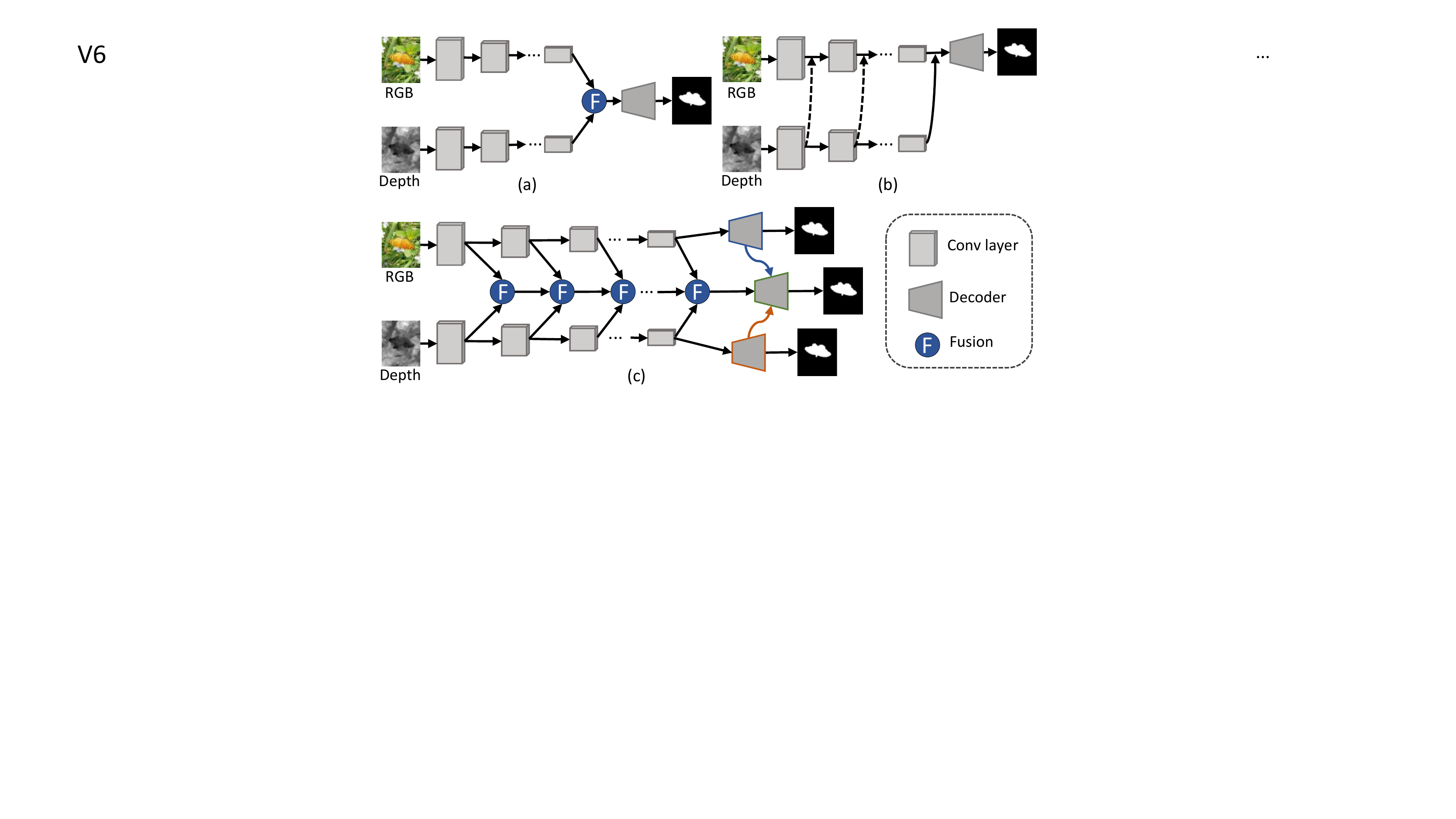}
    \caption{Comparison between existing RGB-D salient object detection frameworks and our proposed model. (a) RGB and depth images are fed into two independent network streams, and then the fused high-level features are fed into a decoder to obtain the predicted saliency maps (\eg, \cite{chen2018progressively,liu2019two,chen2019three,han2017cnns}). (b) Depth features are integrated into the RGB network using an auxiliary subnetwork (\eg, \cite{chen2019multi,zhao2019contrast,zhu2019pdnet,fan2020bbs,zhai2021bifurcated}). (c) Our method adopts two modality-specific networks and a shared learning network to explicitly explore the modality-specific characteristics and shared information, respectively. Then, the features learned from the modality-specific decoders are integrated into the shared decoder to boost the SOD performance.
    }\vspace{-0.45cm}
    \label{fig01}
\end{figure}

Over the past years, various RGB-D SOD methods have been proposed and they often focus on how effectively fuse RGB and depth images. The existing fusion strategies can be divided into three categories, \ie, early fusion, late fusion, and middle fusion. The \textbf{early fusion} strategy often adopts a simple concatenation to integrate the two modalities. For example, these methods \cite{peng2014rgbd,ren2015exploiting,song2017depth,liu2019salient} directly integrate RGB and depth images to form a four-channel input. However, this type of fusion does not consider the distribution gap between the two modalities, which could result in an inaccurate feature fusion. The \textbf{late fusion} strategy is to use two parallel network streams to generate independent saliency maps for RGB and depth data, then the two maps are fused to obtain a final prediction map \cite{guo2016salient,wang2019adaptive,ding2019depth}. However, it is still challenging to capture the complex interactions between the two modalities.

Recent researches mainly focus on the \textbf{middle fusion} strategy, which utilizes two independent networks to learn intermediate features of the two modalities separately, and then the fused features are fed into a subsequent network or decoder (as shown in Fig.~\ref{fig01} (a)). Besides, other methods carry out cross-modal fusion at multiple scales \cite{chen2018attention,chen2018progressively,liu2019two,chen2019three,han2017cnns,ji2021calibrated,huang2021multi}. As a result, the complex correlations can be effectively exploited from the two modalities. Moreover, several methods utilize depth information to enhance RGB features via an a auxiliary subnetwork \cite{chen2019multi,zhao2019contrast,zhu2019pdnet} (as shown in Fig.~\ref{fig01} (b)). For example, Zhao \etal~\cite{zhao2019contrast} introduced a contrast prior into a CNN-based architecture to enhance the depth information, and then the enhanced depth was integrated with RGB features using a fluid pyramid integration module. Zhu \etal~\cite{zhu2019pdnet} utilized an independent subnetwork to extract depth-based features, which were then incorporated into the RGB network. It should be noted that the above methods mainly focus on learning shared representations by fusing them and then use a decoder to generate the final saliency map. What is more, there is no decoder with supervision to guide the depth-based feature learning \cite{zhao2019contrast,zhu2019pdnet}, which may prevent optimal depth features from being obtained. From a multi-modal learning perspective, several works \cite{hu2017sharable,lu2020cross,zhou2019dual,zhou2020hi} have shown that exploring both the shared information and modality-specific characteristics can improve the model performance. However, in the RGB-D SOD community, few methods explicitly exploit modality-specific characteristics.

To alleviate the above issue, in this paper, we propose a novel RGB-D SOD framework, \ie, \textbf{S}pecificity-\textbf{P}reserving \textbf{N}etwork (termed \ours), which can effectively explore the shared information as well as capture modality-specific characteristics to improve the SOD performance. In the proposed \ours, two encoder subnetworks are used to extract multi-scale features for the two modalities (\ie, RGB and depth), and a cross-enhanced integration module (CIM) is proposed to integrate cross-modal features at different feature layers. Then, we use a simple U-Net \cite{ronneberger2015u} structure to construct a modality-specific decoder, in which skip connections between the encoder and decoder layers are used to combine hierarchical features. In this way, we can learn powerful modality-specific features in each independent decoder, which also captures modality-specific characteristics to provide cross-modal complementary. Further, we construct a shared decoder to combine hierarchical features from outputs of the previous CIM via a skip connection. To make full use of the modality-specific features, a multi-modal feature aggregation (MFA) is proposed to integrate them into the shared decoder. Finally, we formulate a unified and end-to-end trainable framework where shared and modality-specific information can be simultaneously exploited to boost the SOD performance.

The main contributions of our paper are summarized as follows:

\begin{itemize}
	\vspace{4pt}
	\item We propose a novel RGB-D salient object detection framework, \ie, \textbf{S}pecificity-\textbf{P}reserving \textbf{N}etwork (termed \ours), which can explore the shared information from RGB and depth images as well as preserve modality-specific characteristics.
	
	\vspace{4pt}
	\item We propose a cross-enhanced integration module (CIM) to integrate the cross-modal features and learn shared representations for the two modalities. More importantly, the output of each CIM is propagated to the next layer to explore rich cross-level information.
	
	\vspace{4pt}
	\item We propose an effective multi-modal feature aggregation (MFA) module to integrate the learned modality-specific features. By using it, our model can make full use of the features learned in the modality-specific decoder to boost the salient object detection performance.
	
	\vspace{4pt}
	\item Extensive experiments on six public RGB-D SOD and three camouflaged object detection (COD) datasets demonstrate the superiority of our model over other cutting-edge methods. Moreover, we carry out an attribute-based evaluation to study the performance of many state-of-the-art RGB-D SOD methods under different challenging factors (\eg, number of salient objects, indoor or outdoor environments, light conditions, and object scale), which has not been done previously by existing studies.
	\vspace{4pt}
\end{itemize}

This paper significantly extends our previous work published in the ICCV-2021 \cite{zhouiccv21}, with multi-fold improvements as follows. (1) We provide some insightful discussions for the differences between the proposed CIM and some existing fusion strategies (see ~\ref{CIM}), and the differences between the proposed CIM and MFA (Sec.~\ref{MFA}). (2) We provide more details to the conference version. Specifically, we add a subsection to review some existing RGB
SOD methods and discuss the importance of integrating multi-level/scale features (refer to Sec.~\ref{sod}). Besides, we provide the details of evaluation metrics to better understand their characteristics (see Sec.~\ref{metrics}).
(3) We provide an additional ablation study and attribute-based evaluation. We validate the effectiveness of the shared decoder (\ref{shared_ablation}), and study the effects on different numbers of CIM (Sec.~\ref{cim_ablation}). Besides, we add an attribute-based evaluation on object scale, and the results also show our model can effectively handle scale variations of the objects (see Sec.~\ref{attribute}).
(4) We extend the proposed \ours~ to a new RGB-D task, \ie, COD. Quantitative and qualitative evaluations conducted on three COD benchmarks demonstrate the superiority of our \ours~over other existing RGB and RGB-D COD methods (presented in Sec.~\ref{COD_enten}).

\section{Related Work}

In this section, we review three types of works that are most related to the proposed model, \ie, RGB salient object detection, RGB-D salient object detection, and multi-modal learning.

\subsection{RGB Salient Object Detection}
\label{sod}

Early salient object detection methods are based on hand-crafted features and some saliency priors, such as background prior \cite{zhu2014saliency}, color contrast \cite{achanta2009frequency}, compactness prior \cite{zhou2015salient}, and center prior \cite{jiang2013submodular}. However, the generalization and effectiveness of these traditional methods are limited. With the breakthrough of deep learning in the field of computer vision, various deep learning-based salient object detection methods have been developed and obtained promising performance. For example, Hou \etal \cite{hou2017deeply} propose a novel salient object detection method by introducing short connections to the skip-layer structures within the holistically-nested edge detector architecture. Wang \etal \cite{wang2018salient} propose a recurrent fully convolutional network framework for salient object detection and it obtains promising performance. Liu \etal \cite{liu2018picanet} propose to hierarchically embed global and local context modules into the top-down pathway, which can generate attention over the context regions for each pixel. Deng \etal \cite{deng2018r3net} propose a recurrent residual refinement network with residual refinement blocks to accurately detect salient objects. More methods can be found in a survey paper \cite{wang2021salient}. Besides, scale variation is one key challenge in the SOD task, thus several methods have been proposed to effectively integrate multi-level/scale features \cite{wang2017edge,zhang2017amulet,zhang2018bi,pang2020multi} to boost the SOD performance. As in the proposed method, we mainly consider how to effectively cross-modal features (\ie, RGB and depth), and the multi-level information can be exploited via the proposed cross-enhanced integration module.

\subsection{RGB-D Salient Object Detection}

Early RGB-D based SOD methods often extract hand-crafted features from the input RGB-D data. For example, Lang \etal~\cite{lang2012depth} proposed the first RGB-D SOD work, which utilized Gaussian mixture models to model the distribution of depth-induced saliency. After that, several methods were explored based on different principles, such as center-surround difference \cite{ju2014depth,guo2016salient}, contrast \cite{desingh2013depth,peng2014rgbd,ren2015exploiting}, center/boundary prior \cite{zhu2017innovative,liang2018stereoscopic}, and background enclosure \cite{feng2016local}. However, these methods usually suffer from unsatisfactory performance due to the limited expression ability of handcrafted features. Benefiting from the rapid development of deep convolutional neural networks (CNNs), several deep learning-based works \cite{qu2017rgbd,zhao2019contrast,piao2019depth,zhang2020uc,fan2019rethinking} have recently been developed and obtained promising results. For example, Qu \etal~\cite{qu2017rgbd} develop a CNN model to fuse saliency cues from different low levels into hierarchical features for boosting the SOD performance. Chen \etal~\cite{chen2018progressively} propose a complementarity-aware fusion module to effectively integrate cross-modal and cross-level features for RGB and depth modalities. Piao \etal~\cite{piao2019depth} propose a depth-induced multi-scale recurrent attention network to enhance the cross-modality feature fusion. Fan \etal~\cite{fan2019rethinking} design a depth depurator unit to filter out some low-quality depth maps. Most other models \cite{chen2018attention,liu2019two,chen2019three,han2017cnns,li2020,lieccv20} employ cross-modal fusion at multiple scales using different integration strategies.

\begin{figure*}
	\begin{centering}
		\includegraphics[width=1.0\textwidth]{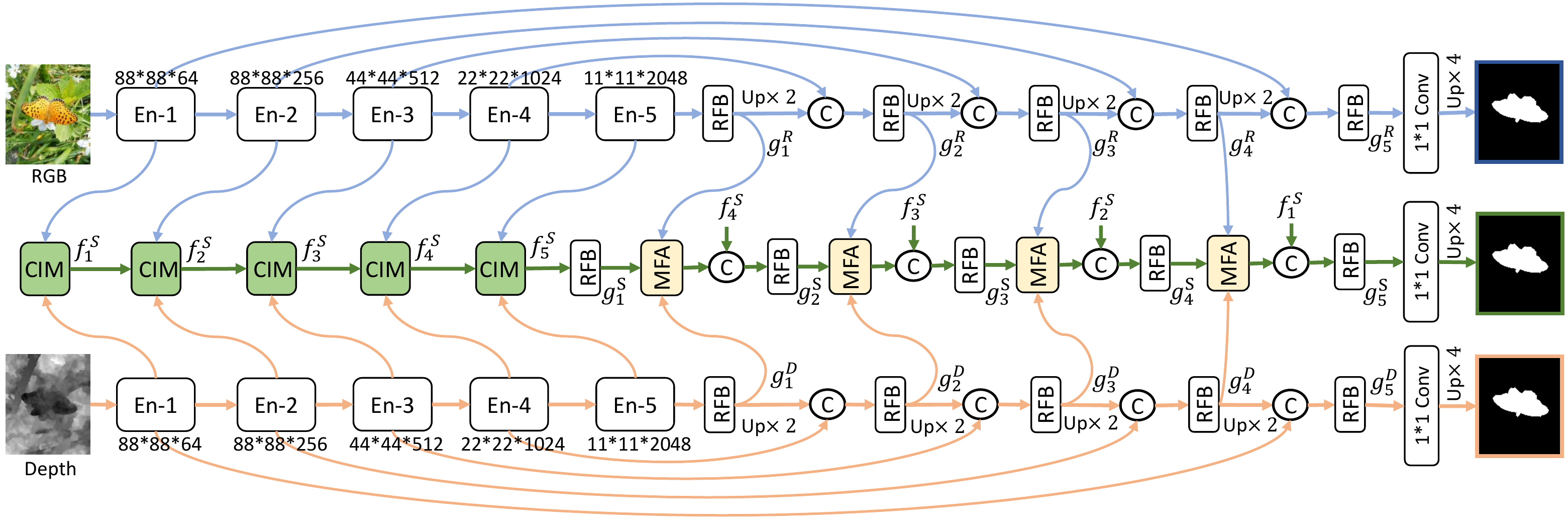}
		\caption{The overall architecture of the proposed \ours. Our model consists of two modality-specific learning networks and a shared learning network. The modality-specific learning networks are used to preserve the individual properties for each modality (\ie, RGB or depth), while the shared network is used to fuse cross-modal features and explore their complementary information. Skip connections are adopted to combine hierarchical features between the encoder and decoder layers. The learned features from the modality-specific decoder are integrated into the shared decoder to provide rich multi-modal complementary information for boosting saliency detection performance. Here, “C” denotes feature concatenation.
		}\vspace{-0.25cm}
		\label{fig02}
	\end{centering}
\end{figure*}

\subsection{Multi-modal Learning}

Recently, multi-modal (or multi-view) learning has attracted more and more attention, as most data can be collected from multiple sources or represented with different types of features. One traditional strategy is to directly concatenate the feature vectors from such multi-modal data into a feature vector. However, this simple concatenation may fail to exploit the complex correlations across multi-modal data. As such, several multi-modal learning methods have been developed to explicitly fuse the complementary information from different modalities to improve model performance. These popular methods can be divided into three following types. 1) Co-training \cite{chaudhuri2009multi,ding2015robust} tries to minimize the disagreement between different modalities, 2) Multiple kernel learning \cite{gonen2011multiple} utilizes a predefined set of kernels from multiple modalities and integrates these modalities using the learned weights of the kernels, and 3) Subspace learning \cite{white2012convex,zhang2017latent} assumes that there exists a latent subspace shared by different modalities, in which multiple modalities can be originated from one underlying latent representation. Besides, to effectively fuse multi-modal data, several deep learning-based models have also been explored. For example, Ngiam \etal~\cite{ngiam2011multimodal} propose to learn a shared representation from audio and video inputs. Eitel \etal~\cite{eitel2015multimodal} adopt two separate CNN streams for RGB and depth, respectively, and then combine them using a late fusion network to achieve RGB-D object recognition. Besides, Hu \etal~\cite{hu2017sharable} present a shareable and individual multi-view learning algorithm to explore more properties of multi-modal data. Lu \etal~\cite{lu2020cross} present a shared-specific feature transfer framework to achieve a cross-modal person ReID task.

\section{Methodology}
In this section, we first present the overall framework of the proposed \ours~ in Sec.~\ref{sec3.1}. Then we describe the two key components in our model, \ie, the modality-specific learning network and shared learning network, in Sec.~\ref{sec3.2} and Sec.~\ref{sec3.3}, respectively. Finally, Sec.~\ref{sec3.4} provides the overall loss function.

\subsection{Overview}
\label{sec3.1}

Fig.~\ref{fig02} shows the framework of the proposed specificity-preserving network for RGB-D SOD. First, the RGB and depth images are fed into two-stream modality-specific learning networks to obtain their multi-level feature representations, and a CIM is proposed to learn their shared feature representation. Second, the specific and shared decoder subnetworks are utilized to generate saliency prediction maps, respectively. Besides, the original features from the encoder networks are integrated into the decoder via a skip connection. Finally, to make full use of the features learned by using the modality-specific decoder, we propose an MFA module to effectively integrate these features into the shared decoder. We give the details of each key part below.

\subsection{Modality-specific Learning Network}
\label{sec3.2}

As shown in Fig.~\ref{fig02}, the modality-specific subnetwork is built using the Res2Net-50 \cite{pami20Res2net}, which has been pretrained on ImageNet \cite{russakovsky2015imagenet} dataset. Thus, there are five multi-level features, \ie, $F^{R}=[f_m^{R},m=1,2,\dots,5]$ and $F^{D}=[f_m^{D},m=1,2,\dots,5]$, in the modality-specific encoder subnetworks for RGB and depth, respectively. In our study, we denote the input resolution of the modality-specific encoder subnetwork as $W\times{H}$. Thus, we have a feature resolution of $\frac{H}{8}*\frac{W}{8}$ for the first layer, and a general resolution of $\frac{H}{2^{m}}*\frac{W}{2^{m}}$ (when $m>1$). Besides, the channel number of features in the $m$-th layer is given as $C_m$ ($m=1,2,\dots$), and we have $C=[64,256,512,1024,2048]$.

After obtaining the high-level features $f_5^{R}$ and $f_5^{D}$, they are then fed into the modality-specific decoder subnetworks to generate individual saliency maps. Besides, we utilize a U-Net \cite{ronneberger2015u} structure to construct the modality-specific decoder, where the skip connections between the encoder and decoder layers are used to combine hierarchical features. Moreover, the concatenated features (only $f_5^{R}$ or $f_5^{D}$ in the first stage of the decoder subnetwork) are fed to the receptive field block (RFB) \cite{wu2019cascaded} to capture global context information. It is worth noting that the modality-specific learning network enables us to learn effective and powerful individual features for each modality by retaining its specific properties. These features are then integrated into the shared decoder subnetwork to boost the saliency detection performance.

\begin{figure}
	\begin{centering}
		\includegraphics[width=0.46\textwidth]{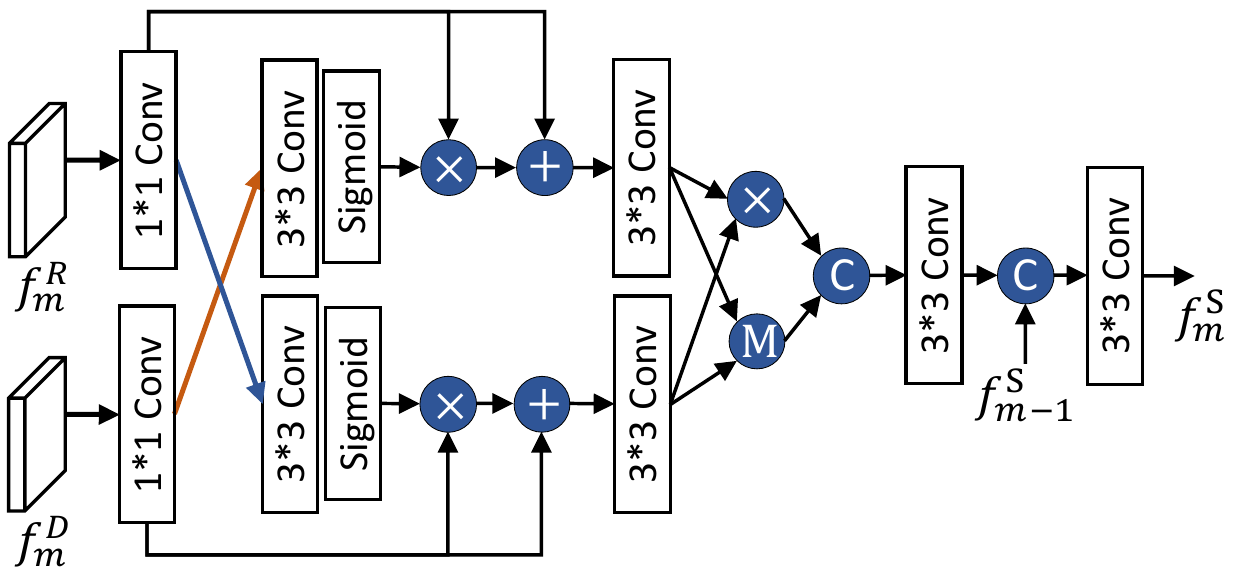}
		\caption{Diagram of the proposed cross-enhanced integration module (CIM).
		Here, ``C" denotes feature concatenation, and ``+", ``$\times$", and ``M" denote the element-wise addition, multiplication, and maximization, respectively.
		}
		\label{fig03}
	\end{centering}
\end{figure}

\subsection{Shared Learning Network}
\label{sec3.3}

As shown in Fig.~\ref{fig02}, in the shared learning network, we fuse the cross-modal features from the RGB and depth modalities to learn their shared representation, which is fed into the shared decoder to generate the final saliency map. Besides, we also adopt skip connections between the encoder and decoder layers to combine hierarchical features. Moreover, to make full use of the features learned by the modality-specific decoder, we integrate them into the shared decoder to improve the saliency detection performance.

\subsubsection{Cross-enhanced Integration Module}
\label{CIM}

We propose a CIM to effectively fuse cross-modal features. Taking $f_{m}^{R}\in\mathbb{R}^{W_m*H_m*C_m}$ and $f_{m}^{D}\in\mathbb{R}^{W_m*H_m*C_m}$ as an example (for convenience, the width, height, and channel number of the $m$-th layer are denoted as $W_m$, $H_m$, and $C_m$),
we use a $1\times1$ convolutional layer to reduce the channel number to $C_m/2$ for acceleration. The CIM includes two parts, \ie, cross-modal feature enhancement and adaptive feature fusion. First, we use a cross-enhanced strategy to exploit the correlations between the two modalities by learning their enhanced features. Specifically, as shown in Fig.~\ref{fig03}, the two features can be fed into a $3\times{3}$ convolutional layer with a \emph{Sigmoid} activation function, and then we can obtain the normalized feature maps, \ie,
$w_m^{R}=\sigma(Conv_3(f_{m}^{R}))\in[0,1]$ and $w_m^{D}=\sigma(Conv_3(f_{m}^{R}))\in[0,1]$,
where $\sigma$ is the logistic \emph{Sigmoid} activation function. To exploit the correlations between the two modalities, the normalized feature maps can be regarded as feature-level attention maps to adaptively enhance the feature representation. In this way, the feature map from one modality can be used to enhance another modality. Besides, to preserve the original information of each modality, a residual connection is adapted to combine the enhanced features with their original features. Thus, we have the cross-enhanced feature representations for the two modalities as follows: 
\begin{equation}
\left\{
\begin{aligned}
f_m^{R'}=f_{m}^{R}+f_{m}^{R}\otimes{w_m^{D}} ,\\
f_m^{D'}=f_{m}^{D}+f_{m}^{D}\otimes{w_m^{R}} ,\\
\end{aligned}
\right.
\end{equation}
where $\otimes$ denotes element-wise multiplication.

\begin{figure}
	\begin{centering}
		\includegraphics[width=0.8\columnwidth]{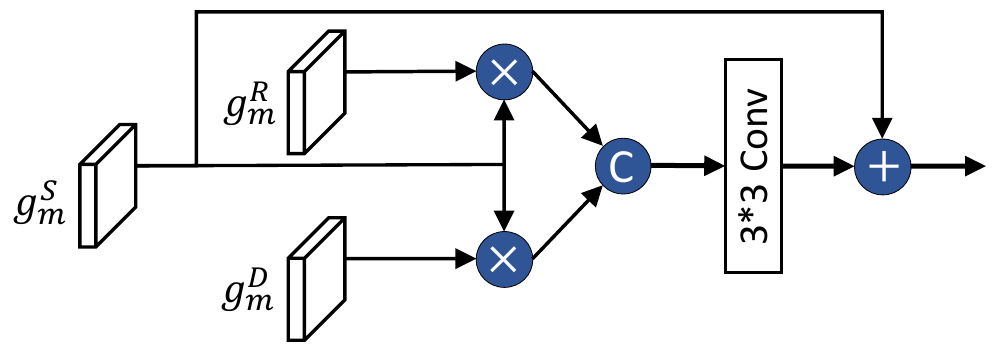}
		\caption{Diagram of the proposed multi-modal feature aggregation (MFA) module. Here ``+", ``$\times$", ``C" denote element-wise addition, element-wise multiplication, and feature concatenation, respectively.}
		\label{fig04}
	\end{centering}
\end{figure}

Once we have obtained the cross-enhanced feature representations (\ie, $f_m^{R'}$ and $f_m^{D'}$), one critical task is to effectively fuse them. Various strategies can be used to fuse features from different modalities, including element-wise multiplication and maximization. However, it is unclear which is best for specific tasks. In order to benefit from the advantages of different strategies, we apply element-wise multiplication and maximization, and then concatenate the results together. Specifically, the two features $f_m^{R'}$ and $f_m^{D'}$ are first fed into a $3\times{3}$ convolutional layer to obtain their smooth representations, and then we carry out element-wise multiplication and maximization. Thus, we can obtain:
\begin{equation}
\left\{
\begin{aligned}
&p_{mul}=Bconv_3(f_m^{R'})\otimes{Bconv_3(f_m^{D'})} ,\\
&p_{max}=Max(Bconv_3(f_m^{R'}),{Bconv_3(f_m^{D'})}) , \\
\end{aligned}
\right.
\end{equation}
where $Bconv(\cdot)$ is a sequential operation that combines a $3\times{3}$ convolution followed by
batch normalization, and a \emph{ReLU} function. Then, we concatenate the results as $p_{cat}=[p_{mul},p_{max}]\in\mathbb{R}^{W_m*H_m*C_m}$, and obtain $p_{cat}^1=Bconv_{3}(p_{cat})$ through a $Bconv_3$ operation to adaptively weigh the two parts. Further, the output $p_{cat}^1$ is concatenated with the previous output $f_{m-1}^{S}$ of the ($m-1$)-th CIM, and fed into the second $Bconv_3$ operation. Finally, we obtain the output $f_m^{S}$ of the $m$-th CIM. Note that, when $m=1$, we do not need to use a $1\times1$ convolutional layer for reducing the channel number. Besides, there is no previous output $f_{m-1}^{S}$ (when $m=1$), so we only feed the concatenated features into one $Bconv_3$ operation.

\textbf{Remarks}. It is worth noting that our CIM can effectively exploit the correlations between the two modalities via cross-enhanced feature learning, and fuse them by adaptively weighting the different feature representations. Besides, the fused feature representation ${f}_m^{S}$ is propagated to the next layer to capture and integrate cross-level information. As some works \cite{liu2019salient,peng2014rgbd,song2017depth} directly integrate RGB images and depth maps to form a four-channel input (\ie, a cascade operation), while other methods carry out cross-modal fusion strategies, \eg, attention-based fusion modules \cite{chen2019three,chen2018attention}, fusion-refinement module (\eg, summation) \cite{liu2019two}, etc. Different from these methods, the proposed CIM mainly exploits the correlation between RGB and depth images, and then adaptively integrates enhanced cross-modal features to obtain their fused feature representation.

\subsubsection{Multi-modal Feature Aggregation}
\label{MFA}

To make full use of the features learned in the modality-specific decoder, we propose a simple but effective MFA module to integrate them into the shared decoder. Specifically, in the $m$-th layer of the shared decoder, we have the shared representation $g_m^S$, and the learned features $g_m^R$ and $g_m^D$ in the modality-specific decoder. As shown in Fig.~\ref{fig04}, two features $g_m^R$ and $g_m^D$ are multiplied by the shared features of the current layer, \ie, $g_m^{RS}=g_m^S\otimes{g_m^R}$ and $g_m^{DS}=g_m^S\otimes{g_m^D}$. The two features are further concatenated ($[g_m^{DR},g_m^{DS}]$) and then fed into a $Bconv(\cdot)$ operation to obtain $g_m^{Sc}$. Finally, we obtain the output of the MFA module to combine the convolutional feature $g_m^{Sc}$ with the original feature $g_m^S$ via an addition operation.

\textbf{Remarks}. In the MFA, the learned modality-specific features are used to enhance the shared features and provide rich and complementary cross-modal information. Specifically, we use the two modality-specific features (\ie, $g_m^R$ and $g_m^D$) to enhance $g_m^S$. More importantly, the modality-specific decoder is given a supervision signal to guide feature learning for the modality-specific property preservation, which benefits the final prediction results when integrating them into the shared decoder. We also note that the differences between CIM and MFA, \ie, the CIM is used to learn the fused multi-modal (\ie, RGB and depth) feature representation, while the MFA utilizes the learned modality-specific feature to aggregate the feature representation in the shared decoder.

\subsection{Loss Function}
\label{sec3.4}

Finally, we formulate a unified and end-to-end trainable framework. The overall loss function consists of two parts, \ie, $\mathcal{L}_{sp}$ and $\mathcal{L}_{sh}$, for the modality-specific and shared decoders, respectively. For convenience, $S_{R}$ and $S_{D}$ denote the prediction maps when using RGB and depth images, respectively, $S_{sh}$ denotes the prediction map using their shared representation, and $G$ denotes the ground truth map. Therefore, the overall loss function can be formulated as follows:
\begin{equation}
\begin{aligned}
\mathcal{L}_{total}=\mathcal{L}_{sh}(S_{sh},G)+\mathcal{L}_{sp}(S_R,G)+\mathcal{L}_{sp}(S_D,G).
\end{aligned}
\label{eq3}
\end{equation}

In Eq.~(\ref{eq3}), we utilize the pixel position-aware loss \cite{wei2019f3net} for $\mathcal{L}_{sp}$ and $\mathcal{L}_{sh}$, which can pay different attention to hard and easy pixels to boost the performance.

\section{Experimental Results and Analysis}

In this section, we first provide the experimental setup (\ref{setup}), including datasets, evaluation metrics, and implementation details. Then we show the performance comparison (\ref{results}) including quantitative and qualitative evaluation, as well as conduct ablation studies to validate the effectiveness of each key component (\ref{ablation}). Finally, we conduct an attribute-based evaluation to show the effectiveness of our model in dealing with different challenges (\ref{attribute}).

\subsection{Experimental Setup}
\label{setup}

\subsubsection{Datasets}
To validate the effectiveness of the proposed model, we evaluate it on six public RGB-D SOD datasets, including NJU2K \cite{ju2014depth}, NLPR \cite{peng2014rgbd}, DES \cite{cheng2014depth}, SSD \cite{zhu2017three}, STERE \cite{niu2012leveraging} and SIP \cite{fan2019rethinking}. The details of each dataset can be found at: \url{https://github.com/taozh2017/RGBD-SODsurvey}.

For a fair comparison, we utilize the same protocol to form the training and test sets, as introduced in \cite{piao2019depth,fan2019rethinking}. The training set includes 2,195 samples in total, where 1,485 samples from NJU2K \cite{ju2014depth} and 700 samples from the NLPR \cite{peng2014rgbd}. The remaining samples from NJU2K (500) and NLPR (300), and the whole DES (135), SSD (80), STERE (1,000), and SIP (929), are used for testing.

\begin{table*}[t!]
  \centering
  \renewcommand{\arraystretch}{1.25}
  \renewcommand{\tabcolsep}{0.3mm}
  \caption{
  Benchmarking results of 8 representative traditional models and 23 deep models on six public RGB-D saliency detection datasets using four widely used evaluation metrics (\ie, $S_{\alpha}$ \cite{fan2017structure}, max $E_{\phi}$\cite{Fan2018Enhanced}, max $F_{\beta}$ \cite{achanta2009frequency}, and $\mathcal{M}$ \cite{perazzi2012saliency}).
  ``$\uparrow$`` \& ``$\downarrow$'' indicate that larger or smaller is better. The subscript of each model denotes the publication year. The best results are highlighted in \textbf{Bold} fonts.} 

  \scriptsize
  \begin{tabular}{r||p{0.6cm}p{0.6cm}p{0.6cm}p{0.6cm}|p{0.6cm}p{0.6cm}p{0.6cm}p{0.6cm}|p{0.6cm}p{0.6cm}p{0.6cm}p{0.6cm}|p{0.56cm}p{0.56cm}p{0.56cm}p{0.56cm}|p{0.56cm}p{0.56cm}p{0.56cm}p{0.56cm}|p{0.56cm}p{0.56cm}p{0.56cm}p{0.56cm}}

  \hline \toprule

    &\multicolumn{4}{c|}{NJU2K~\cite{ju2014depth}}
    &\multicolumn{4}{c|}{STERE~\cite{niu2012leveraging}}
    &\multicolumn{4}{c|}{DES~\cite{cheng2014depth}}
    &\multicolumn{4}{c|}{NLPR~\cite{peng2014rgbd}}
    &\multicolumn{4}{c|}{SSD~\cite{zhu2017three}}
    &\multicolumn{4}{c}{SIP~\cite{fan2019rethinking}}\\

    \textbf{Model}
    &$S_{\alpha}\uparrow$   &$F_{\beta}\uparrow$    &$E_{\xi}\uparrow$  &$\mathcal{M}\downarrow$
    &$S_{\alpha}\uparrow$   &$F_{\beta}\uparrow$    &$E_{\xi}\uparrow$  &$\mathcal{M}\downarrow$
    &$S_{\alpha}\uparrow$   &$F_{\beta}\uparrow$    &$E_{\xi}\uparrow$  &$\mathcal{M}\downarrow$
    &$S_{\alpha}\uparrow$   &$F_{\beta}\uparrow$    &$E_{\xi}\uparrow$  &$\mathcal{M}\downarrow$
    &$S_{\alpha}\uparrow$   &$F_{\beta}\uparrow$    &$E_{\xi}\uparrow$  &$\mathcal{M}\downarrow$
    &$S_{\alpha}\uparrow$   &$F_{\beta}\uparrow$    &$E_{\xi}\uparrow$  &$\mathcal{M}\downarrow$ \\

    \midrule

 LHM$_{14}$~\cite{peng2014rgbd}
    & .514   & .632   & .724   & .205
    & .562   & .683   & .771   & .172
    & .562   & .511   & .653   & .114
    & .630   & .622   & .766   & .108
    & .566   & .568   & .717   & .195
    & .511   & .574   & .716   & .184	\\

    ACSD$_{14}$~\cite{ju2014depth}
    & .699   & .711   & .803   & .202
    & .692   & .669   & .806   & .200
    & .728   & .756   & .850   & .169
    & .673   & .607   & .780   & .179
    & .675   & .682   & .785   & .203
    & .732   & .763   & .838   & .172	\\

    LBE$_{16}$~\cite{feng2016local}
    & .695   & .748   & .803   & .153
    & .660   & .633   & .787   & .250
    & .703   & .788   & .890   &.208
    & .762   & .745   & .855   & .081
    & .621   & .619   & .736   & .278
    & .727   & .751   & .853   & .200	\\

    DCMC$_{16}$~\cite{cong2016saliency}
    & .686   & .715   & .799   & .172
    & .731   & .740   & .819   & .148
    & .707   & .666   & .773   & .111
    & .724   & .648   & .793   & .117
    & .704   & .711   & .786   & .169
    & .683   & .618   & .743   & .186	\\

    SE$_{16}$~\cite{guo2016salient}
    & .664   & .748   & .813   & .169
    & .708   & .755   & .846   & .143
    & .741   &.741    &.856    & .090
    & .756   & .713   & .847   & .091
    & .675   & .710   & .800   & .165
    & .628   & .661   & .771   & .164	\\

    MDSF$_{17}$~\cite{song2017depth}
    & .748   & .775   & .838   & .157
    & .728   & .719   & .809   & .176
    &  .741  & .746   & .851   & .122
    & .805   & .793   & .885   & .095
    & .673   & .703   & .779   & .192
    & .717   & .698   & .798   & .167	\\

    CDCP$_{17}$~\cite{zhu2017innovative}
    & .669 & .621 & .741 & .180
    & .713 & .664 & .786 & .149
    & .709 & .631 & .811 & .115
    & .669 & .621 & .741 & .180
    & .603 & .535 & .700 & .214
    & .595 & .505 & .721 & .224 \\

    DTM$_{20}$~\cite{cong2019going}
    & .706 & .716 & .799 & .190
    & .747 & .743 & .837 & .168
    & .752 & .697 & .858 & .123
    & .733 & .677 & .833 & .145
    & .677 & .651 & .773 & .199
    & .690 & .659 & .778 & .203 \\

    \midrule
    DF$_{17}$~\cite{qu2017rgbd}
    & .763 & .804 & .864 & .141
    & .757 & .757 & .847 & .141
    & .752 & .766 & .870 & .093
    & .802 & .778 & .880 & .085
    & .747 & .735 & .828 & .142
    & .653 & .657 & .759 & .185 \\

    CTMF$_{18}$~\cite{han2017cnns}
    & .849 & .845 & .913 & .085
    & .848 & .831 & .912 & .086
    & .863 & .844 & .932 & .055
    & .860 & .825 & .929 & .056
    & .776 & .729 & .865 & .099
    & .716 & .694 & .829 & .139 \\

    PCF$_{18}$~\cite{chen2018progressively}
    & .877   & .872   & .924   & .059
    & .875   & .860   & .925   & .064
    & .842   & .804   & .893   & .049
    & .874   & .841   & .925   & .044
    & .841   & .807   & .894   & .062
    & .842   & .838   & .901   & .071	\\

    {AFNet}$_{19}$~\cite{wang2019adaptive}
    & .772   & .775   & .853   & .100
    & .825   & .823   & .887   & .075
    & .770   & .729   & .881   & .068
    & .799   & .771   & .879   & .058
    & .714   & .687   & .807   & .118
    & .720   & .712   & .819   & .118	\\

    CPFP$_{19}$~\cite{zhao2019contrast}
    & .878   & .877   & .923   & .053
    & .879   & .874   & .925   & .051
    & .872   & .846   & .923   & .038
    & .888   & .867   & .932   & .036
    & .807   & .766   & .852   & .082
    & .850   & .851   & .903   & .064	\\

    MMCI$_{19}$~\cite{chen2019multi}
    & .859   & .853   & .915   & .079
    & .873   & .863   & .927   & .068
    & .848   & .822   & .928   & .065
    & .856   & .815   & .913   & .059
    & .813   & .781   & .882   & .082
    & .833   & .818   & .897   & .086	\\

    TANet$_{19}$~\cite{chen2019three}
    & .878   & .874   & .925   & .060
    & .871   & .861   & .923   & .060
    & .858   & .827   & .910   & .046
    & .886   & .863   & .941   & .041
    & .839   & .810   & .897   & .063
    & .835   & .830   & .895   & .075	\\

    DMRA$_{19}$~\cite{piao2019depth}
    & .886   & .886   & .927   & .051
    & .886   & .886   & .938   & .047
    & .900   & .888   & .943   & .030
    & .899   & .879   & .947   & .031
    & .857   & .844   & .906   & .058
    & .806   & .821   & .875   & .085	\\

    cmSalGAN$_{20}$~\cite{jiang2020cmsalgan}
    & .903   & .896   & .940   & .046
    & .900   & .894   & .936   & .050
    & .913   & .899   & .943   & .028
    & .922   & .907   & .957   & .027
    & .791   & .735   & .867   & .086
    & .865   & .864   & .906   & .064	\\

    ASIFNet$_{20}$~\cite{li2020asif}
    & .889   & .888   & .927   & .047
    & .878   & .878   & .927   & .049
    & .934   & .935   & .974   & .019
    & .906   & .888   & .944   & .030
    & .857   & .834   & .884   & .056
    & .857   & .859   & .896   & .061	\\

    ICNet$_{20}$~\cite{li2020icnet}
    & .894   & .891   & .926   & .052
    & .903   & .898   & .942   & .045
    & .920   & .913   & .960   & .027
    & .923   & .908   & .952   & .028
    & .848   & .841   & .902   & .064
    & .854   & .857   & .903   & .069	\\

    A2dele$_{20}$~\cite{piao2020}
    & .871   & .874   & .916   & .051
    & .878   & .879   & .928   & .044
    & .886   & .872   & .920   & .029
    & .898   & .882   & .944   & .029
    & .802   & .776   & .861   & .070
    & .828   & .833   & .889   & .070	\\

    JL-DCF$_{20}$~\cite{fu2020jl}
    & .903   & .903   & .944   & .043
    & .905   & .901   & .946   & .042
    & .929   & .919   & .968   & .022
    & .925   & .916   & {.962}   & .022
    & .830   & .795   & .885   & .068
    & .879   & .885   & .923   & .051	\\

    S${^2}$MA$_{20}$~\cite{liu2020}
    & .894   & .889   & .930   & .053
    & .890   & .882   & .932   & .051
    & .941   & .935   & .973   & .021
    & .915   & .902   & .953   & .030
    & .868   & .848   & .909   & .052
    & .872   & .877   & .919   & .057	\\

    UCNet$_{20}$~\cite{zhang2020uc}
    & .897   & .895   & .936   & .043
    & .903   & .899   & .944   & .039
    & .933   & .930   & .976   & .018
    & .920   & .903   & .956   & .025
    & .865   & .854   & .907   & .049
    & .875   & .879   & .919   & .051	\\

    SSF$_{20}$~\cite{zhang2020}
    & .899   & .896   & .935   & .043
    & .893   & .890   & .936   & .044
    & .904   & .884   & .941   & .026
    & .914   & .896   & .953   & .026
    & .845   & .824   & .897   & .058
    & .876   & .882   & .922   & .052	\\

    HDFNet$_{20}$~\cite{paneccv2020}
    & .908   & .911   & .944   & .038
    & .900   & .900   & .943   & .041
    & .926   & .921   & .970   & .021
    & .923   & .917   & \textbf{.963}   & .023
    & \textbf{.879}   & .870   & \textbf{.925}   & .045
    & .886   & .894   & .930   & .047	\\

    Cas-GNN$_{20}$~\cite{luoECCV2020}
    & .911   & .903   & .933   & .035
    & .899   & .901   & .930   & .039
    & .905   & .906   & .947   & .028
    & .919   & .904   & .947   & .028
    & .872   & .862   & .915   & .047
    & .875   & .879   & .919   & .051	\\

    CMMS$_{20}$~\cite{li2020}
    & .900   & .897   & .936   & .044
    & .895   & .893   & .939   & .043
    & .937   & .930   & .976   & .018
    & .915   & .896   & .949   & .027
    & .874   & .864   & .922   & .046
    & .872   & .877   & .911   & .058	\\

    CoNet$_{20}$~\cite{Wei_2020_ECCV}
    & .895   & .893   & .937   & .046
    & \textbf{.908}   & .905   & \textbf{.949}   & .040
    & .909   & .896   & .945   & .028
    & .908   & .887   & .945   & .031
    & .853   & .840   & .915   & .059
    & .858   & .867   & .913   & .063	\\

    DANet$_{20}$~\cite{zhaoeccv20}
    & .899   & .910   & .935   & .045
    & .901   & .892   & .937   & .043
    & .924   & .928   & .968   & .023
    & .915   & .916   & .953   & .028
    & .864   & .866   & .914   & .050
    & .875   & .892   & .918   & .054	\\

    PGAR$_{20}$~\cite{chen2020progressively}
    & .909   & .907   & .940   & .042
    & .907   & .898   & .939   & .041
    & .913   & .902   & .945   & .026
    & \textbf{.930}   & .916   & .961   & .024
    & .865   & .838   & .898   & .057
    & .876   & .876   & .915   & .055	\\

    D$^{3}$Net$_{21}$~\cite{fan2019rethinking}
    & .900   & .900   & .950   & .041
    & .899   & .891   & .938   & .046
    & .898   & .885   & .946   & .031
    & .912   & .897   & .953   & .030
    & .857   & .834   & .910   & .058
    & .860   & .861   & .909   & .063	\\

    \midrule

    \ours~(Ours)
    & \textbf{.925} & \textbf{.935} & \textbf{.954} & \textbf{.028}
    & {.907} & \textbf{.915} & {.944} & \textbf{.037}
    & \textbf{.945} & \textbf{.950} & \textbf{.980} & \textbf{.014}
    & {.927} & \textbf{.925} & {.959} & \textbf{.021}
    & .871 & \textbf{.883} & .915 & \textbf{.044}
    & \textbf{.894} & \textbf{.916} & \textbf{.930} & \textbf{.043} \\

  \bottomrule
  \hline
  \end{tabular}\label{tab1}
\end{table*}

\subsubsection{Evaluation Metrics}
\label{metrics}

We adopt four widely used metrics to evaluate the effectiveness of the proposed model. The detailed definitions of the four metrics are provided as follows.
\begin{itemize}{\setlength{\parsep}{-0.25ex}}

\item {\textbf{Precision-recall (PR)}} \cite{achanta2009frequency}. Given a saliency map $S$, we convert it to a binary map $M$, and then we can compute the $\textup{Precision}$
and $\textup{Recall}$ by
\begin{equation}
\textup{Precision}=\frac{|M\cap{G}|}{|M|},~ \textup{Recall}=\frac{|M\cap{G}|}{|G|},
\end{equation}
where $G$ denotes the ground-truth. A popular strategy is to partition $S$ by using a set of thresholds (\ie, varying from $0$ to $255$). For each threshold, we calculate a pair of recall and precision scores, and then combine all scores to obtain a PR curve.

\item \textbf{Structure Measure}. {S-measure} ($S_{\alpha}$) \cite{fan2017structure} is proposed to assess the structural similarity between the regional perception ($S_r$) and object perception ($S_o$), which is defined by
\begin{equation}
\begin{aligned}
S_{\alpha}=\alpha * S_{o}+\left(1-\alpha\right)*S_{r},
\end{aligned}
\end{equation}
where $ \alpha \in \left[ 0,1\right]$ is a trade-off parameter and it is set to $0.5$ as default \cite{fan2017structure}.

\item \textbf{Enhanced-alignment Measure}. $E_{\phi}$ \cite{Fan2018Enhanced} is used to capture image-level statistics and their local pixel matching information, and it is defined as
\begin{equation}
\begin{aligned}
E_{\phi}=\frac{1}{W*H}\sum_{i=1}^{W}\sum_{i=1}^{H}\phi_{FM}\left(i,j\right),
\end{aligned}
\end{equation}
where $ \phi_{FM} $ denotes the enhanced-alignment matrix \cite{Fan2018Enhanced}.

\item \textbf{F-measure} ($F_{\beta}$ \cite{achanta2009frequency}). It is used to comprehensively consider both precision and recall, and we can obtain the weighted harmonic mean by
\begin{equation}
\begin{aligned}
F_{\beta}=\left(1+\beta ^2\right)\frac{\textup{Precision}*\textup{Recall}}{\beta^{2}\textup{Precision}+\textup{Recall}},
\end{aligned}
\end{equation}
where $\beta^2$ is set to $0.3$ to emphasize the precision \cite{achanta2009frequency}. We use different fixed $[0,255]$ thresholds to compute the $F$-measure. This yields a set of $F$-measure values for which we report the maximal $F_{\beta}$ in our experiments.

\item \textbf{Mean Absolute Error} ($\mathcal{M}$). It is adopted to evaluate the average pixel-level relative error between the ground truth (\ie, $G$) and normalized prediction (\ie, $S$), which is defined by
\begin{equation}
\mathcal{M}=\frac{1}{W*H}\sum_{i=1}^{W}\sum_{i=1}^{H}\left|S\left(i,j\right)-G\left(i,j\right) \right|,
\end{equation}
where $W$ and $H$ denote the width and height of the map, respectively. $\mathcal{M}$ estimates the similarity between the saliency map and the ground-truth map, and normalizes it to $[0,1]$.\\

\end{itemize}

\subsubsection{Implementation Details}

The proposed model is implemented with the PyTorch library, and trained on one NVIDIA Tesla V100 GPU with 32 GB memory. The backbone network (Res2Net-50 \cite{pami20Res2net}) is used, which has been pre-trained on ImageNet \cite{russakovsky2015imagenet}. When using the backbone network, since RGB and depth images have different channels, the input channel of the depth encoder is modified to $1$. We utilize the Adam algorithm to optimize the proposed model. The initial learning rate is set to $1e-4$ and is divided by $10$ every $60$ epochs. The input resolutions of RGB and depth images are resized to $352\times{352}$. To enhance the generalizability of the proposed learning algorithm, we adopt multiple data augmented strategies, consisting of random flipping, rotating, and border clipping. The batch size is set to $20$ and the model has trained over $200$ epochs.

For testing, the RGB and depth images are first resized to $352\times{352}$ and then fed into the model to obtain the predicted saliency map. Then, the predicted saliency map is resized back to the original size of the input images. Finally, the output of the shared decoder can be regarded as the final prediction for our model.

\subsection{Performance Comparison}
\label{results}

\subsubsection{Compared RGB-D SOD Models}

We compare the proposed \ours~with 30 benchmarking RGB saliency detection methods, including 8 handcrafted traditional models (\ie, LHM \cite{peng2014rgbd}, ACSD \cite{ju2014depth}, LBE \cite{feng2016local}, DCMC \cite{cong2016saliency}, SE \cite{guo2016salient}, MDSF \cite{song2017depth}, CDCP \cite{zhu2017innovative}), and DTM~\cite{cong2019going}, and 23 deep models (\ie, DF \cite{qu2017rgbd}, CTMF~\cite{han2017cnns}, PCF \cite{chen2018progressively}, AFNet \cite{wang2019adaptive}, CPFP \cite{zhao2019contrast}, MMCI \cite{chen2019multi}, TANet \cite{chen2019three}, DMRA \cite{piao2019depth}, cmSalGAN \cite{jiang2020cmsalgan}, ASIFNet \cite{li2020asif}, ICNet \cite{li2020icnet}, A2dele \cite{piao2020}, JL-DCF \cite{fu2020jl}, S$^2$MA \cite{liu2020}, UCNet \cite{zhang2020uc}, SSF \cite{zhang2020}, HDFNet~\cite{paneccv2020}, Cas-GNN \cite{luoECCV2020}, CMMS \cite{li2020}, D$^3$Net \cite{fan2019rethinking}, CoNet \cite{Wei_2020_ECCV}, DANet \cite{zhaoeccv20}, and PGAR \cite{chen2020progressively}). Details for these methods can be referred to the related papers and the survey paper \cite{zhou2021rgb}.

\begin{figure*}
	\begin{centering}
		\includegraphics[width=1.0\textwidth]{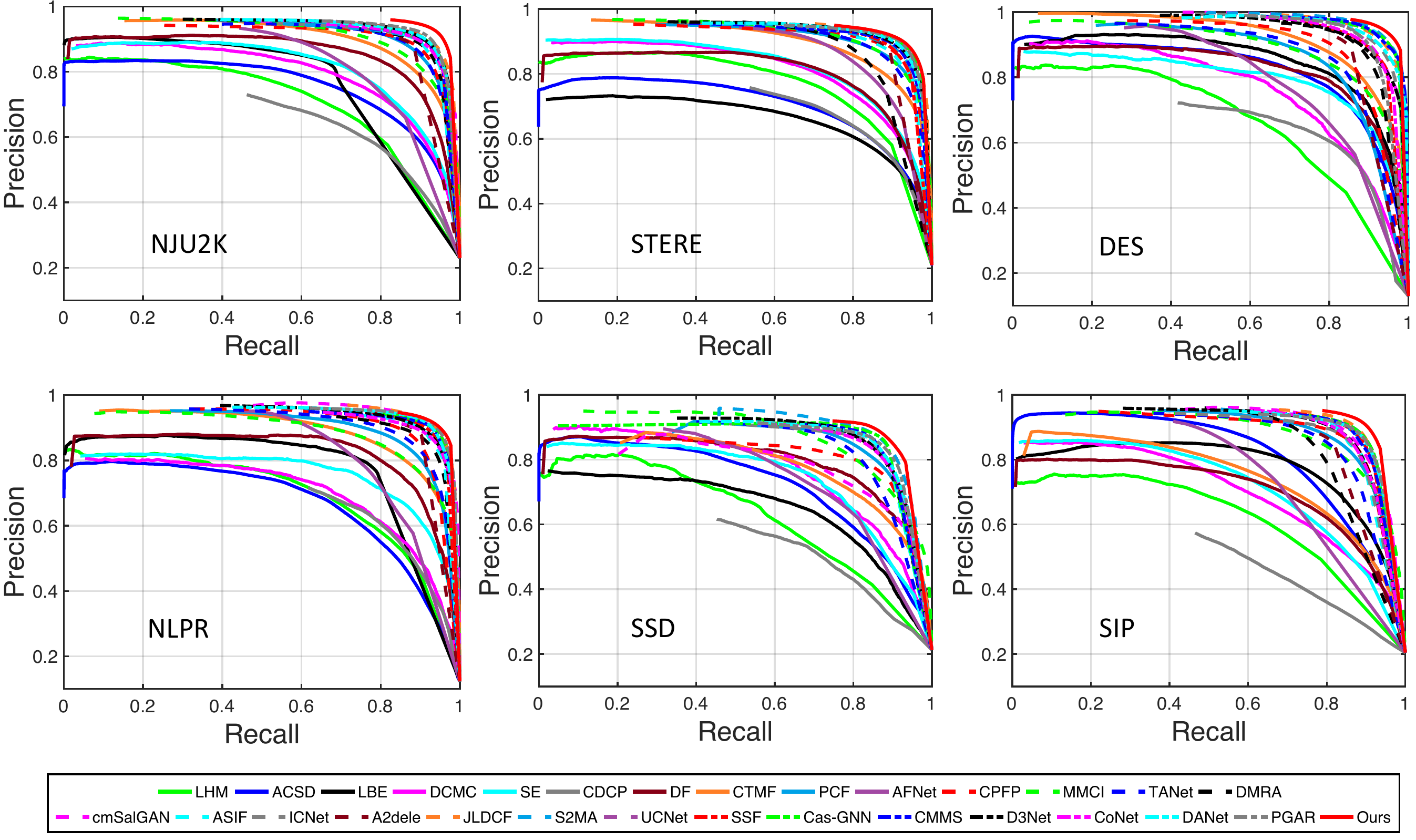}
		\caption{PR curves on six datasets (\ie, NJU2K \cite{ju2014depth}, STERE \cite{niu2012leveraging}, DES \cite{cheng2014depth}, NLPR \cite{peng2014rgbd}, SSD~\cite{zhu2017three}, and SIP \cite{fan2019rethinking}).}
		\label{fig05}
	\end{centering}
\end{figure*}

\begin{figure*}
	\begin{centering}
		\includegraphics[width=1.0\textwidth]{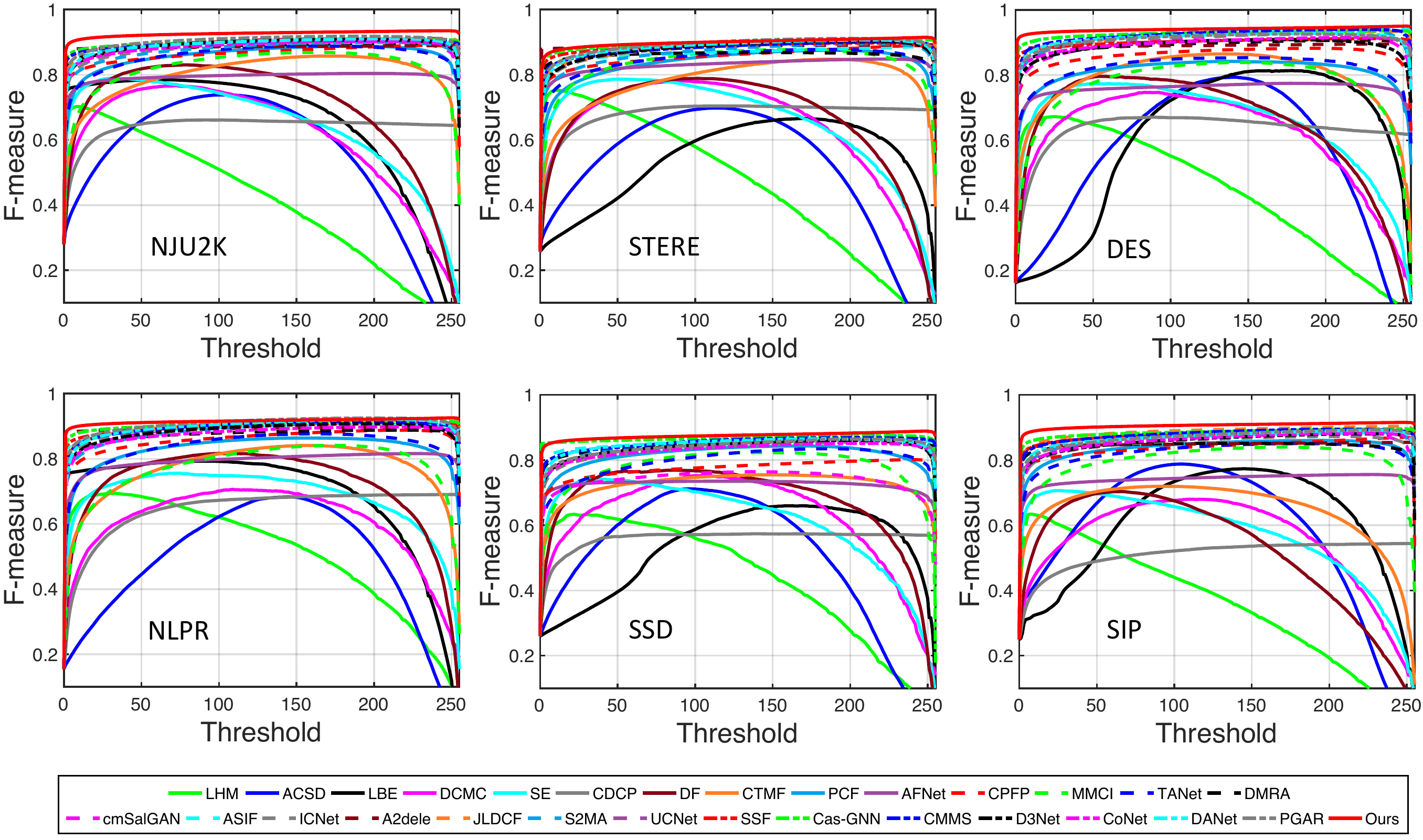}
		\caption{F-measure curves under different thresholds on six datasets (\ie, NJU2K \cite{ju2014depth}, STERE \cite{niu2012leveraging}, DES \cite{cheng2014depth}, NLPR \cite{peng2014rgbd}, SSD~\cite{zhu2017three}, and SIP \cite{fan2019rethinking}).}
		\label{fig06}
	\end{centering}
\end{figure*}

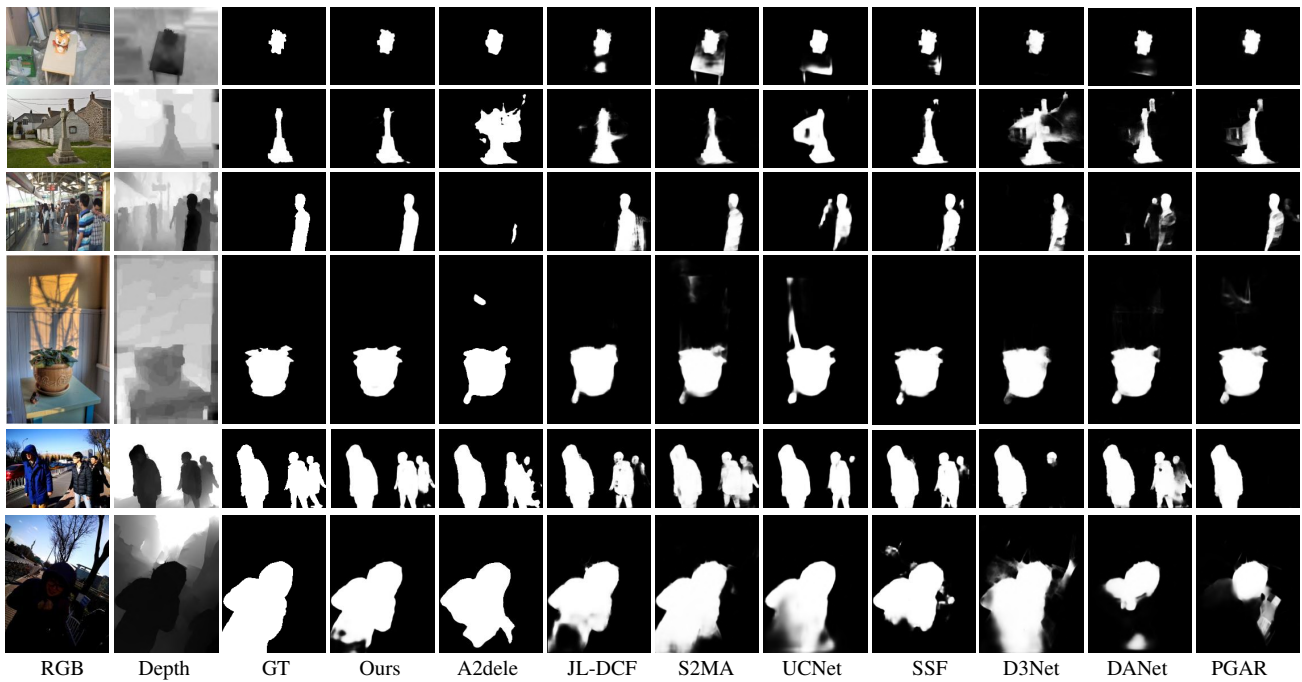
\begin{figure*}[t]
\begin{overpic}[width=0.99\linewidth]{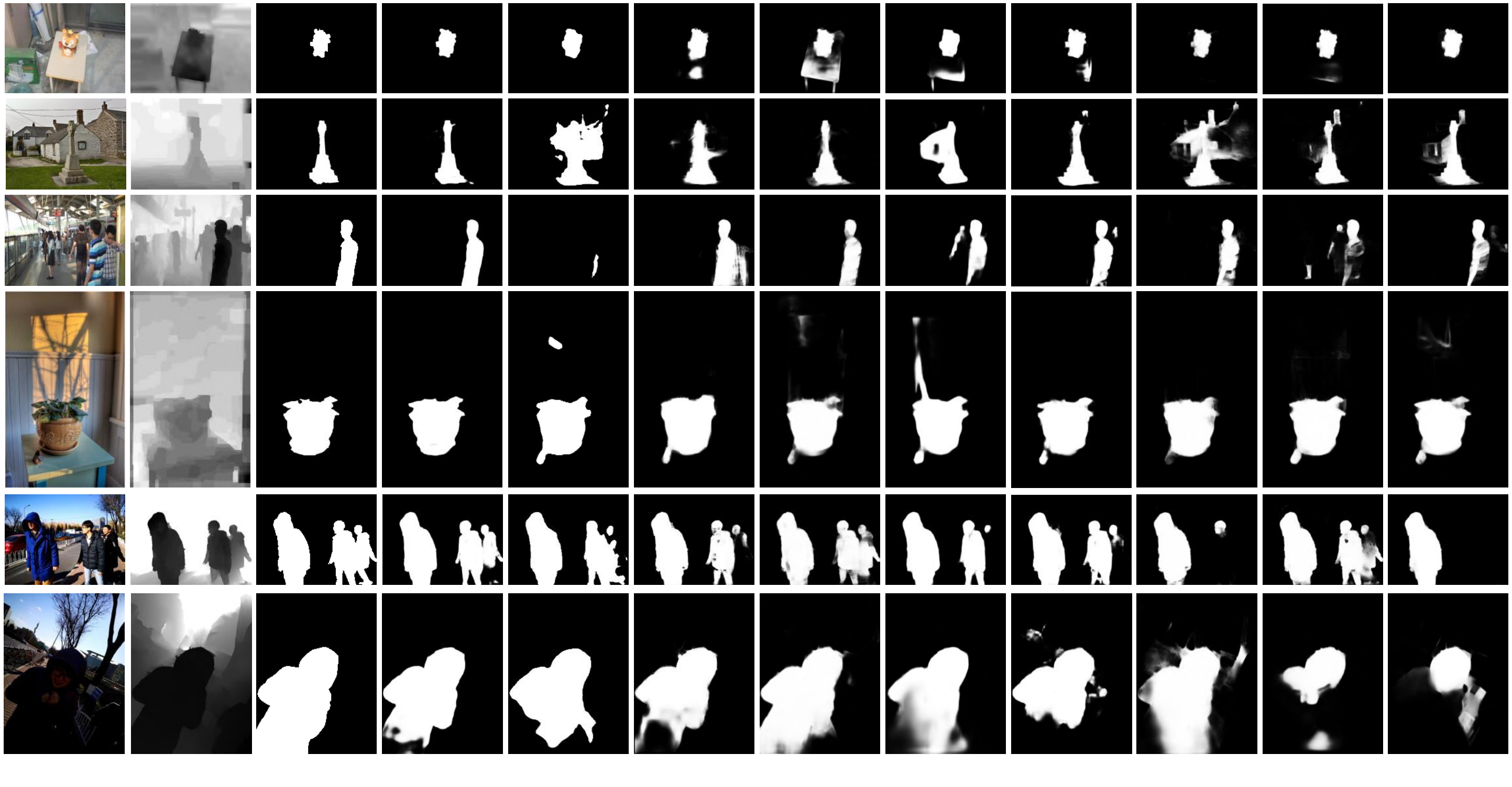}
\put(3, 0.5){\footnotesize RGB}
\put(10.5, 0.5){\footnotesize Depth}
\put(20,0.5){\footnotesize GT}
\put(27.5,0.5){\footnotesize Ours}
\put(35,0.5){\footnotesize A2dele}
\put(43.5,0.5){\footnotesize JL-DCF}
\put(52,0.5){\footnotesize S2MA}
\put(60,0.5){\footnotesize UCNet}
\put(70,0.5){\footnotesize SSF}
\put(77,0.5){\footnotesize D3Net}
\put(85,0.5){\footnotesize DANet}
\put(93,0.5){\footnotesize PGAR}
\end{overpic}
    \caption{Visual comparisons of our method and eight state-of-the-art methods (including A2dele \cite{piao2020}, JL-DCF \cite{fu2020jl}, S2MA \cite{liu2020}, UCNet \cite{zhang2020uc}, SSF \cite{zhang2020}, D3Net \cite{fan2019rethinking}, DANet \cite{zhaoeccv20}, and PGAR \cite{chen2020progressively}.}
    \label{fig_07}
\end{figure*}

\begin{table*}[t!]
  \centering
  \renewcommand{\arraystretch}{1.2}
  \renewcommand{\tabcolsep}{1.99mm}
  \caption{Comparison results of our model and 13 state-of-the art methods (\ie, CTMFF~\cite{han2017cnns}, PCF~ \cite{chen2018progressively}, AFNet \cite{wang2019adaptive}, MMCI \cite{chen2019multi}, CPFP \cite{zhao2019contrast}, DMRA \cite{piao2019depth}, TANet \cite{chen2019three}, A2dele \cite{piao2020}, UCNet \cite{zhang2020uc}, JL-DCF \cite{fu2020jl}, S$^2$MA \cite{liu2020}, SSF \cite{zhang2020}, and D$^3$Net \cite{fan2019rethinking}) on the ReDWeb-S dataset.} 
  \scriptsize
  \begin{tabular}{r|c|c|c|c|c|c|c|c|c|c|c|c|c|c}
  \whline{1pt}

    \textbf{Models}
    & CTMF & PCF & AFNet   & MMCI  & CPFP   & DMRA & TANet   & A2dele   & UCNet   & JL-DCF & S$^2$MA   & SSF  & D$^3$Net & Ours\\

    \whline{1pt}

    $S_{\alpha}\uparrow$
    & 0.641    & 0.655    & 0.546  & 0.660      & 0.685    & 0.592  & 0.656 & 0.641    & 0.713  & 0.734   & 0.711 &0.595 & 0.689 & 0.710\\

    $F_{\beta}\uparrow$
    & 0.607    & 0.627    & 0.549  & 0.641      & 0.645    & 0.579  & 0.623 & 0.603    & 0.710  & 0.727   & 0.696 &0.558 & 0.673 & 0.715\\

    $E_{\phi}\uparrow$
    & 0.739    & 0.743    & 0.693  & 0.754      & 0.744    & 0.721  & 0.741 & 0.672    & 0.794  & 0.805   & 0.781 &0.710 & 0.768 & 0.800\\

    $\mathcal{M}\downarrow$
    & 0.204    & 0.166    & 0.213  & 0.176      & 0.142    & 0.188  & 0.165 & 0.160    & 0.130  & 0.128   & 0.139 &0.189 & 0.149 & 0.129\\

\whline{1pt}

  \end{tabular}\label{tab2}
\end{table*}

\begin{table*}[t!]
  \centering
  \renewcommand{\arraystretch}{1.25}
  \renewcommand{\tabcolsep}{0.2mm}
  \caption{Comparison of our model using different backbone networks.} 

  \scriptsize
  \begin{tabular}{r||p{0.6cm}p{0.6cm}p{0.6cm}p{0.6cm}|p{0.6cm}p{0.6cm}p{0.6cm}p{0.6cm}|p{0.6cm}p{0.6cm}p{0.6cm}p{0.6cm}|p{0.56cm}p{0.56cm}p{0.56cm}p{0.56cm}|p{0.56cm}p{0.56cm}p{0.56cm}p{0.56cm}|p{0.56cm}p{0.56cm}p{0.56cm}p{0.56cm}}
  \hline \toprule

    &\multicolumn{4}{c|}{NJU2K~\cite{ju2014depth}}
    &\multicolumn{4}{c|}{STERE~\cite{niu2012leveraging}}
    &\multicolumn{4}{c|}{DES~\cite{cheng2014depth}}
    &\multicolumn{4}{c|}{NLPR~\cite{peng2014rgbd}}
    &\multicolumn{4}{c|}{SSD~\cite{zhu2017three}}
    &\multicolumn{4}{c}{SIP~\cite{fan2019rethinking}}\\

    \textbf{Model}
    &$S_{\alpha}\uparrow$   &$F_{\beta}\uparrow$    &$E_{\xi}\uparrow$  &$\mathcal{M}\downarrow$
    &$S_{\alpha}\uparrow$   &$F_{\beta}\uparrow$    &$E_{\xi}\uparrow$  &$\mathcal{M}\downarrow$
    &$S_{\alpha}\uparrow$   &$F_{\beta}\uparrow$    &$E_{\xi}\uparrow$  &$\mathcal{M}\downarrow$
    &$S_{\alpha}\uparrow$   &$F_{\beta}\uparrow$    &$E_{\xi}\uparrow$  &$\mathcal{M}\downarrow$
    &$S_{\alpha}\uparrow$   &$F_{\beta}\uparrow$    &$E_{\xi}\uparrow$  &$\mathcal{M}\downarrow$
    &$S_{\alpha}\uparrow$   &$F_{\beta}\uparrow$    &$E_{\xi}\uparrow$  &$\mathcal{M}\downarrow$ \\

  \midrule

    Ours(ResNet-50)
    & .922 & .934 & .952 & .030
    & .904 & .914 & .942 & .037
    & .936 & .944 & .974 & .016
    & {.930} & .931 & {.965} & .020
    & .869 & .876 & .906 & .044
    & {.896} & .916 & .934 & .041 \\

    Ours(Res2Net-50)
    & {.925} & {.935} & {.954} & {.028}
    & {.907} & {.915} & {.944} & {.037}
    & {.945} & {.950} & {.980} & {.014}
    & {.927} & {.925} & {.959} & {.021}
    & .871 & {.883} & .915 & {.044}
    & {.894} & {.916} & {.930} & {.043} \\

  \bottomrule
  \hline
  \end{tabular}\label{tab3}
\end{table*}

\subsubsection{Quantitative Evaluation}

As shown in Table~\ref{tab1}, our method is superior to eight traditional methods (\ie, LHM \cite{peng2014rgbd}, ACSD \cite{ju2014depth}, LBE \cite{feng2016local}, DCMC \cite{cong2016saliency}, SE \cite{guo2016salient}, MDSF \cite{song2017depth}, and CDCP \cite{zhu2017innovative}) by a large margin on all six datasets. Besides, our method outperforms all of the comparison state-of-the-art methods and obtains the best performance in terms of four evaluation metrics on NJU2K, DES, and SIP datasets. Moreover, it is worth noting that our model obtains better performance on STERE and NLPR than most compared RGB-D saliency detection methods. Our model is also comparable with CoNet on the STERE dataset, and JL-DCF and PGAR on the NLPR dataset. Overall, our proposed \ours~obtains promising performance in locating salient object(s) in a given scene. In addition, we show the PR curves \cite{borji2015salient} (Fig.~\ref{fig05}) and F-measure curves in Fig.~\ref{fig06}. For a clear view, we provide the results of 29 RGB-D saliency detection methods, including 28 SOTA models with complete saliency maps. As observed, the superiority of our model is more visible on these reported datasets.

In addition, we compare the proposed \ours~with 13 recent state-of-the-art models on the ReDWeb-S dataset. The results of all compared methods are collected from \url{https://github.com/nnizhang/SMAC}, and the results of our method are obtained by testing the model (trained using NJU2K \cite{ju2014depth} and NLPR \cite{peng2014rgbd}) on the ReDWeb-S dataset. The comparison results are shown in Table~\ref{tab2}. From the results, it can be observed that our method performs better than most compared methods, and it is comparable with UCNet and JL-DCF on the ReDWeb-S dataset.

Moreover, we compare the proposed model using different backbone networks, and the results are shown in Table~\ref{tab3}. From the results, we can see that the proposed model obtains better performance when using Res2Net-50 as the backbone, and the model using ResNet-50 as backbone still performs better than other compared methods (see Table~\ref{tab1}).

\subsubsection{Qualitative Evaluation}

Fig.~\ref{fig_07} shows several representative samples of results comparing our model with eight top state-of-the-art methods. The first row shows a scene with a small object. Our method, A2dele, PGAR, and D3Net can accurately detect the salient object, while JL-DCF, S2MA, SSF, and UCNet predict some non-object regions. In the $2^{nd}$ and $3^{rd}$ rows, we show two examples when the scene is with complex backgrounds. From the comparison results, it can be observed that our method and S2MA produce reliable results, while other RGB-D saliency detection models fail to locate the object or confuse the background as a salient object. In the $4^{th}$ row, the comparison methods (except D3Net) locate a non-salient and small object. In the $5^{th}$ row, we show an example with multiple salient objects, where it is challenging to accurately locate all salient objects. Our method locates all salient objects and segments them more accurately, generating sharper edges compared to other approaches. We show an example under low-light conditions in the last row. It can be seen that some approaches fail to detect the entire extent of the salient object. Our model can produce promising results by suppressing background distractors to boost the saliency detection performance.

\begin{table}[t!]
  \centering
  \renewcommand{\arraystretch}{1.2}
  \renewcommand{\tabcolsep}{1.7mm}
  \caption{Comparisons with inference time and model size of different methods.} 
  \scriptsize
  \begin{tabular}{r|c|c|c}
  \whline{1pt}

    \textbf{Method}
    & Ours   & JL-DCF~\cite{fu2020jl}  & S2MA~\cite{liu2020} \\

    \textbf{Model Size (MB)}
    & 175.3   & 124.5  & 82.7  \\

    \textbf{Inference Time (ms)}
    & 91.7    & 21.8    & 22.1 \\

\whline{1pt}

    \textbf{Method}
    & UCNet~\cite{zhang2020uc}   & SSF~\cite{zhang2020}  & HDFNet~\cite{paneccv2020} \\

    \textbf{Model Size (MB)}
    & 31.3  & 32.9  & 153.2  \\

    \textbf{Inference Time (ms)}
    & 31.8   & 45.7    & 57.1  \\
  \whline{1pt}

  \end{tabular}\label{tab4}
\end{table}

\subsubsection{Inference Time and Model Size}

We test the inference time for different methods on NVIDIA TESLA P40 GPU with 24G memory. The inference time and model size of different methods (including our \ours, JL-DCF~\cite{fu2020jl}, S2MA~\cite{liu2020}, UCNet~\cite{zhang2020uc}, SSF~\cite{zhang2020}, and  HDFNet~\cite{paneccv2020} ) are shown in Table~\ref{tab4}. Because our model adopts two modality-specific networks and a shared learning network to generate individual and shared saliency prediction maps, it has a relatively large model size and takes much inference time for the saliency prediction than other compared methods. Thus, we can design lightweight networks to improve the efficiency of the proposed \ours~in future work.

\begin{table}[t!]
  \centering
  \renewcommand{\arraystretch}{1.2}
  \renewcommand{\tabcolsep}{0.3mm}
  \caption{Quantitative evaluation for ablation studies.
  } 

  \scriptsize
  \begin{tabular}{c|cc|cc|cc|cc|cc|cc}
  \hline\toprule

    &\multicolumn{2}{c|}{NJU2K~\cite{ju2014depth}}
    &\multicolumn{2}{c|}{STERE~\cite{niu2012leveraging}}
    &\multicolumn{2}{c|}{DES~\cite{cheng2014depth}}
    &\multicolumn{2}{c|}{NLPR~\cite{peng2014rgbd}}
    &\multicolumn{2}{c|}{SSD~\cite{zhu2017three}}
    &\multicolumn{2}{c}{SIP~\cite{fan2019rethinking}}\\

    &$S_{\alpha}\uparrow$   &$\mathcal{M}\downarrow$
    &$S_{\alpha}\uparrow$    &$\mathcal{M}\downarrow$
    &$S_{\alpha}\uparrow$    &$\mathcal{M}\downarrow$
    &$S_{\alpha}\uparrow$    &$\mathcal{M}\downarrow$
    &$S_{\alpha}\uparrow$    &$\mathcal{M}\downarrow$
    &$S_{\alpha}\uparrow$    &$\mathcal{M}\downarrow$ \\

  \midrule
    \textbf{Ours}
    & \textbf{.925}   & \textbf{.028}
    & \textbf{.907}   & \textbf{.037}
    & \textbf{.945}   & \textbf{.014}
    & \textbf{.927}   & \textbf{.021}
    & \textbf{.871}   & \textbf{.044}
    & \textbf{.894}   & \textbf{.043} 	\\

   \midrule

    A1
    & .916   & .034
    & .898   & .042
    & .939   & .016
    & .926   & .022
    & .869   & .047
    & .892   & .044 	\\

    A2
    & .921   & .031
    & .895   & .042
    & .938   & .016
    & .925   & .022
    & .865   & .051
    & .896   & .042 	\\

    A3
    & .919   & .032
    & .895   & .043
    & .938   & .016
    & .929   & .020
    & .864   & .049
    & .887   & .048 	\\

    A4
    & .924   & .029
    & .903   & .038
    & .930   & .019
    & .927   & .023
    & .867   & .049
    & .888   & .046 	\\

    \midrule

    B1
    & .918   & .034
    & .901   & .041
    & .939   & .017
    & .922   & .024
    & .858   & .050
    & .885   & .048 	\\

    B2
    & .924   & .029
    & .900   & .041
    & .941   & .015
    & .926   & .022
    & .864   & .049
    & .893   & .044 	\\

    B3
    & .921   & .031
    & .903   & .039
    & .938   & .016
    & .925   & .022
    & .863   & .050
    & .891   & .045 	\\

    \midrule

    C1
    & .913   & .037
    & .900   & .047
    & .935   & .019
    & .922   & .025
    & .861   & .055
    & .880   & .051 	\\

    C2
    & .916   & .034
    & .906   & .040
    & .923   & .021
    & .924   & .022
    & .866   & .049
    & .882   & .051 	\\

  \bottomrule
  \hline
  \end{tabular}\label{tab5}
\end{table}

\subsection{Ablation Studies}
\label{ablation}

To verify the relative importance of different key components of our model, we conduct ablation studies by removing or replacing them from our full model.

\begin{figure}
	\begin{centering}
		\includegraphics[width=0.5\textwidth]{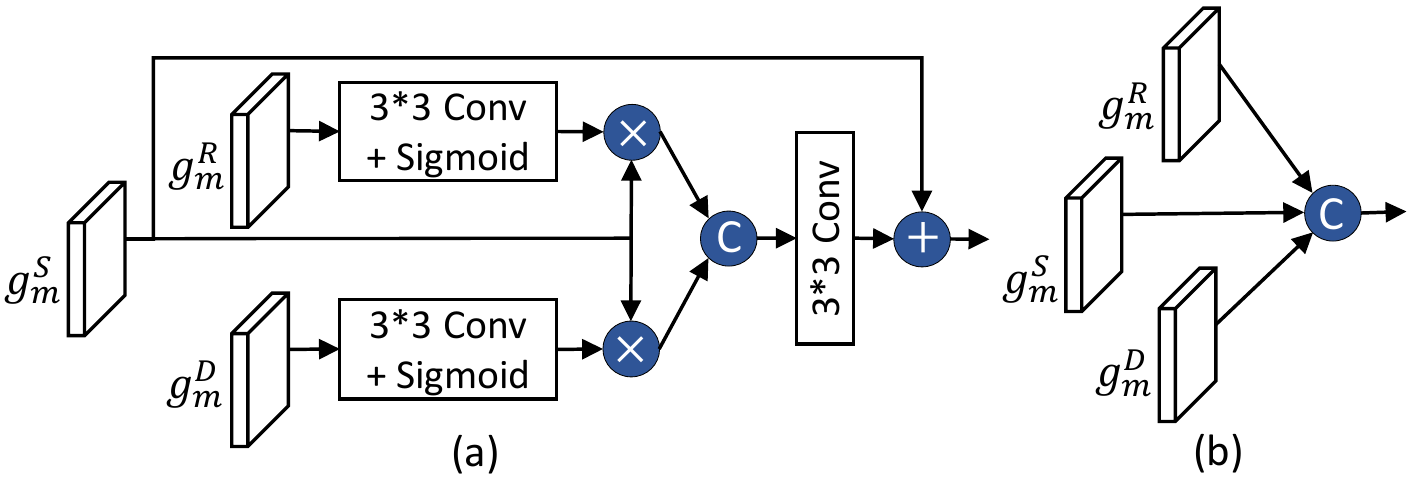}
		\caption{Comparison of MFA module with other fusion strategies.}\vspace{-0.4cm}
		\label{fig08}
	\end{centering}
\end{figure}

\begin{figure*}[!t]
	\begin{centering}
		\includegraphics[width=0.95\textwidth]{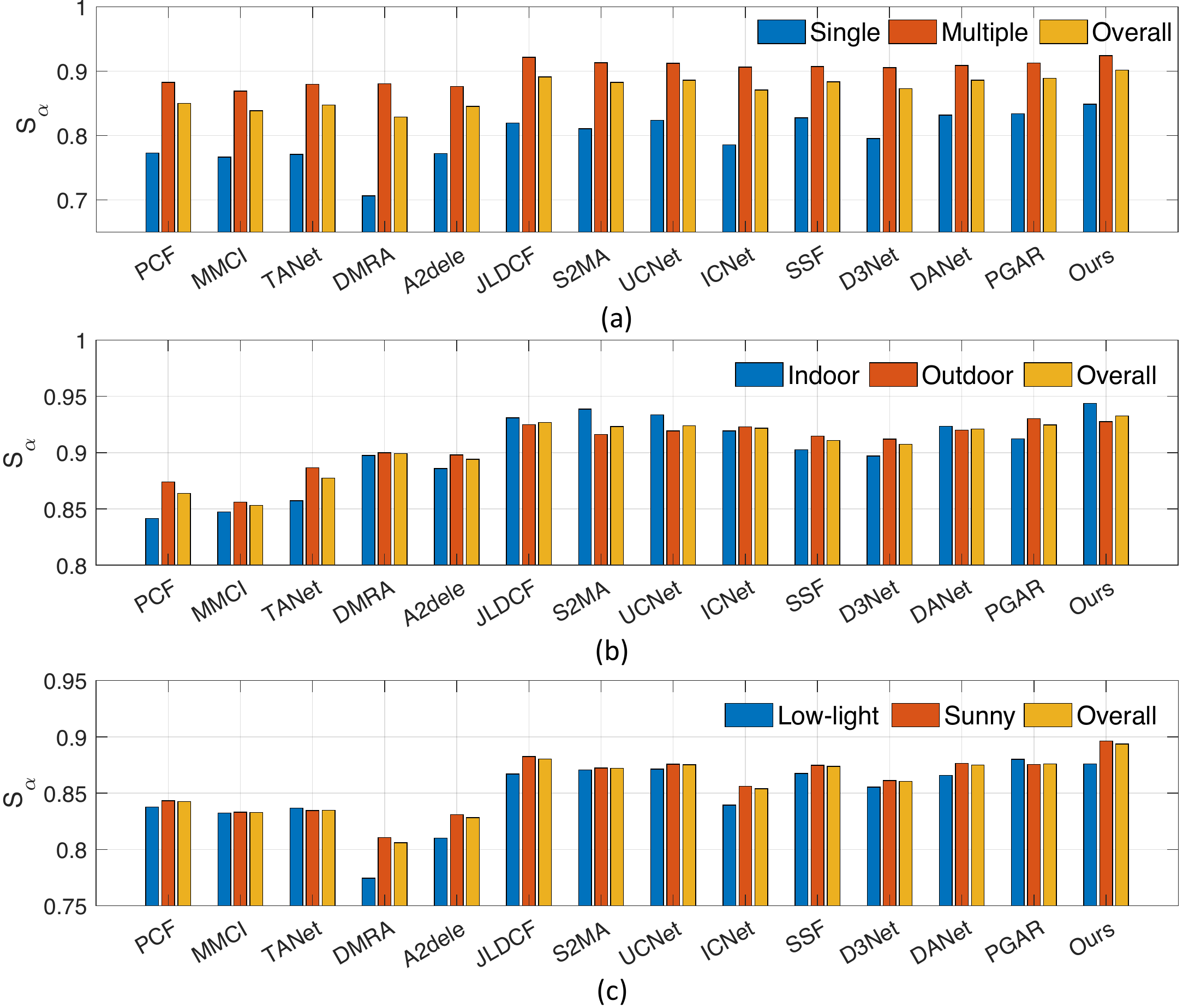}
		\caption{Attribute-based evaluation \emph{w.r.t.} (a) number of salient objects (\ie, single vs. multiple), (b) indoor vs. outdoor environments, and (c) light conditions (low-light vs. sunny). }\vspace{-0.35cm}
		\label{fig09}
	\end{centering}
\end{figure*}

\subsubsection{Effectiveness of CIM}

Since the proposed CIM is used to fuse cross-modal features and learn their shared representation, we utilize a direct concatenation strategy instead of the CIM. Specifically, the two features $f_m^R$ and $f_m^D$ (as shown in Fig.~\ref{fig03}) are directly concatenated and then fed into a $3\times{3}$ convolutional layer to obtain the fused representation in each layer. We denote this evaluation as ``A1" in Table \ref{tab5}. From the comparison results, it can be seen that our model performs better when using the proposed CIM than using a simple feature concatenation strategy. This also indicates the contribution of the CIM in boosting the saliency detection performance. Besides, there are two parts in CIM, \ie, cross-modal feature enhancement and adaptive feature fusion. Thus, to evaluate the contribution of each part, we denote the CIM with only cross-modal feature enhancement or adaptive feature fusion as ``A2" and ``A3", respectively. When comparing the two independent parts with the full version of the CIM, we can see that the effectiveness of the proposed CIM. Moreover, in CIM, the features of the last layer are propagated to the next layer to capture cross-level correlations. To validate the effectiveness of the propagation strategy, we delete this propagation in the CIM, denoted as ``A4". The comparison results between ``A4" and CIM show that this propagation strategy improves the saliency detection performance.

\subsubsection{Effectiveness of MFA}
\label{shared_ablation}

In the proposed framework, the MFA is proposed to make full use of the features learned in the modality-specific decoder, which are then integrated into the shared decoder to provide more multi-modal complementary information. To validate its effectiveness, we delete this module, denoted as ``B1“. Besides, we consider comparing two other feature fusion strategies with our MFA. As shown in Fig.~\ref{fig08}, one is the cross-modal feature enhancement fusion; the other is a simple concatenation strategy. The comparison experiments for the two strategies are denoted ``B2" and ``B3". As shown in Table~\ref{tab5}, Comparing ``B1" and our full model, the results demonstrate the effectiveness of integrating the features learned into the shared decoder. Comparing ``B2" and ``B3" with our full model, we can see that the MFA module outperforms both of the other fusion strategies.

\subsubsection{Effectiveness of Modality-specific Decoder}

We delete the two modality-specific decoders, and the evaluation is as shown in ``C1" of Table~\ref{tab5}. It can be observed that the performance will degrade without using the two parts. This indicates the effectiveness of the modality-specific decoder, which can provide supervision signals to ensure that modality-specific properties can be learned.

Besides, to evaluate the effectiveness of the combination of the two modality-specific decoders, we add an experiment to compare the SOD results when using the output from the shared decoder and the combination of two modality-specific decoders. The evaluation is as shown in ``C2" of Table~\ref{tab5}. From the compared results, we can see that the performance of the shared decoder outperforms the combination of two modality-specific decoders. This indicates the shared decoder can combine multi-modal shared information and modality-specific characteristics to improve the SOD performance.

\begin{table}[t!]
  \centering
  \renewcommand{\arraystretch}{1.2}
  \renewcommand{\tabcolsep}{0.5mm}
  \caption{Ablation study on different numbers of CIM.
  } 
  \scriptsize
  \begin{tabular}{c|cc|cc|cc|cc|cc|cc}
  \hline

    &\multicolumn{2}{c|}{NJU2K}
    &\multicolumn{2}{c|}{STERE}
    &\multicolumn{2}{c|}{DES}
    &\multicolumn{2}{c|}{NLPR}
    &\multicolumn{2}{c|}{SSD}
    &\multicolumn{2}{c}{SIP}\\

    &$S_{\alpha}\uparrow$   &$\mathcal{M}\downarrow$
    &$S_{\alpha}\uparrow$    &$\mathcal{M}\downarrow$
    &$S_{\alpha}\uparrow$    &$\mathcal{M}\downarrow$
    &$S_{\alpha}\uparrow$    &$\mathcal{M}\downarrow$
    &$S_{\alpha}\uparrow$    &$\mathcal{M}\downarrow$
    &$S_{\alpha}\uparrow$    &$M\downarrow$ \\

    \hline

    $\textup{CIM}_1$
    & .918   & .034
    & .908   & .039
    & .929   & .019
    & .928   & .022
    & .865   & .047
    & .889   & .046 	\\

    $\textup{CIM}_3$
    & .920   & .032
    & .900   & .041
    & .935   & .017
    & .928   & .021
    & .857   & .049
    & .891   & .045 	\\

    Ours
    & \textbf{.925}   & \textbf{.028}
    & \textbf{.907}   & \textbf{.037}
    & \textbf{.945}   & \textbf{.014}
    & \textbf{.927}   & \textbf{.021}
    & \textbf{.871}   & \textbf{.044}
    & \textbf{.894}   & \textbf{.043} 	\\

  \hline
  \end{tabular}\label{tab6}
\end{table}

\begin{figure*}
	\begin{centering}
		\includegraphics[width=0.95\textwidth]{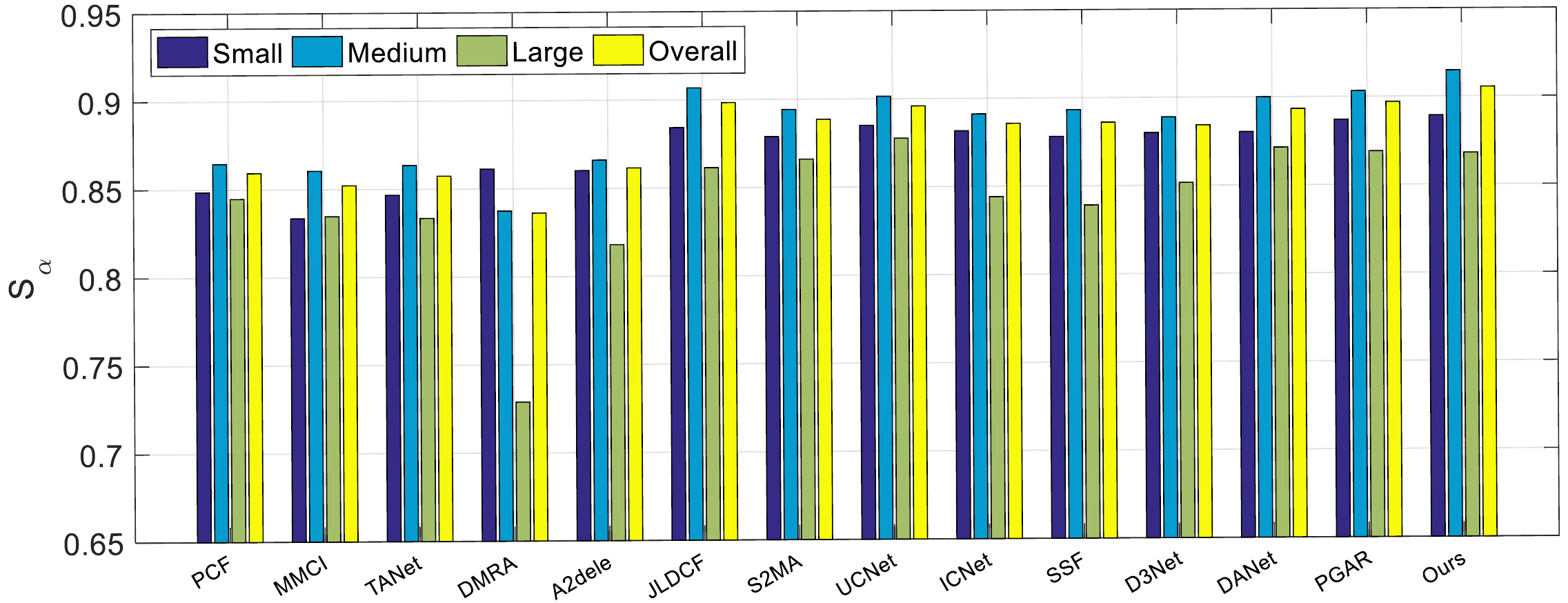}
		\caption{Attribute-based evaluation \emph{w.r.t.} scales of the salient object(s).}\vspace{-0.35cm}
		\label{fig010}
	\end{centering}
\end{figure*}

\subsubsection{Effects on Different Numbers of CIM}
\label{cim_ablation}

To investigate the effects of different numbers of CIM, we compare our full model (\ie, using five CIMs) with two degraded versions, including 1) ``$\textup{CIM}_1$": we only conduct CIM on the features from the last layer in the encoder network; 2) ``$\textup{CIM}_3$": we conduct CIMs on the features from the last three layers in the encoder network, \ie, using three CIMs. Table~\ref{tab6} shows the comparison results using different numbers of CIMs. From the results, we can see that our model with five CIMs obtains better performance on most datasets.

\subsection{Attribute-based Evaluation}
\label{attribute}

There are several challenging factors that affect the performance of RGB-D saliency detection models, such as the number of salient objects, indoor or outdoor environment, light conditions, and so on. Thus, it is interesting to evaluate the saliency detection performance under different conditions, to show the strengths and weaknesses of state-of-the-art models in handling these challenges.

1) \textbf{Single vs. Multiple Objects}. In this evaluation, we construct a hybrid dataset with 1,229 images collected from the NLPR \cite{peng2014rgbd} and SIP \cite{fan2019rethinking} datasets. The comparison results using $S_{\alpha}$ are shown in Fig.~\ref{fig09} (a). As can be observed, it is easier to detect a single salient object than multiple. Besides, our model outperforms other state-of-the-art methods in locating single and multiple objects.

2) \textbf{Indoor vs. Outdoor}. We evaluate the performance of different RGB-D SOD models under indoor and outdoor scenes. In this evaluation experiment, DES \cite{cheng2014depth} and NLPR \cite{peng2014rgbd} include indoor and outdoor scenes, we thus construct a hybrid dataset collected from the two datasets. The comparison results are shown in Fig.~\ref{fig09} (b). As can be observed, many models struggle more to detect salient objects in indoor scenes than outdoor scenes, while JL-DCF, S2MA, UCNet, ICNet, SSF, DANet, and our model perform a little better in outdoor scenes.

3) \textbf{Light Conditions}. We carry out this evaluation on the SIP dataset \cite{fan2019rethinking}, and the data is grouped into two categories, \ie, sunny and low-light. The comparison results are shown in Fig.~\ref{fig09} (c). As can be seen, all models struggle more to detect salient objects in low-light conditions, confirming that low-light negatively impacts SOD performance.

\begin{figure}[t!]
	\begin{centering} \includegraphics[width=0.49\textwidth]{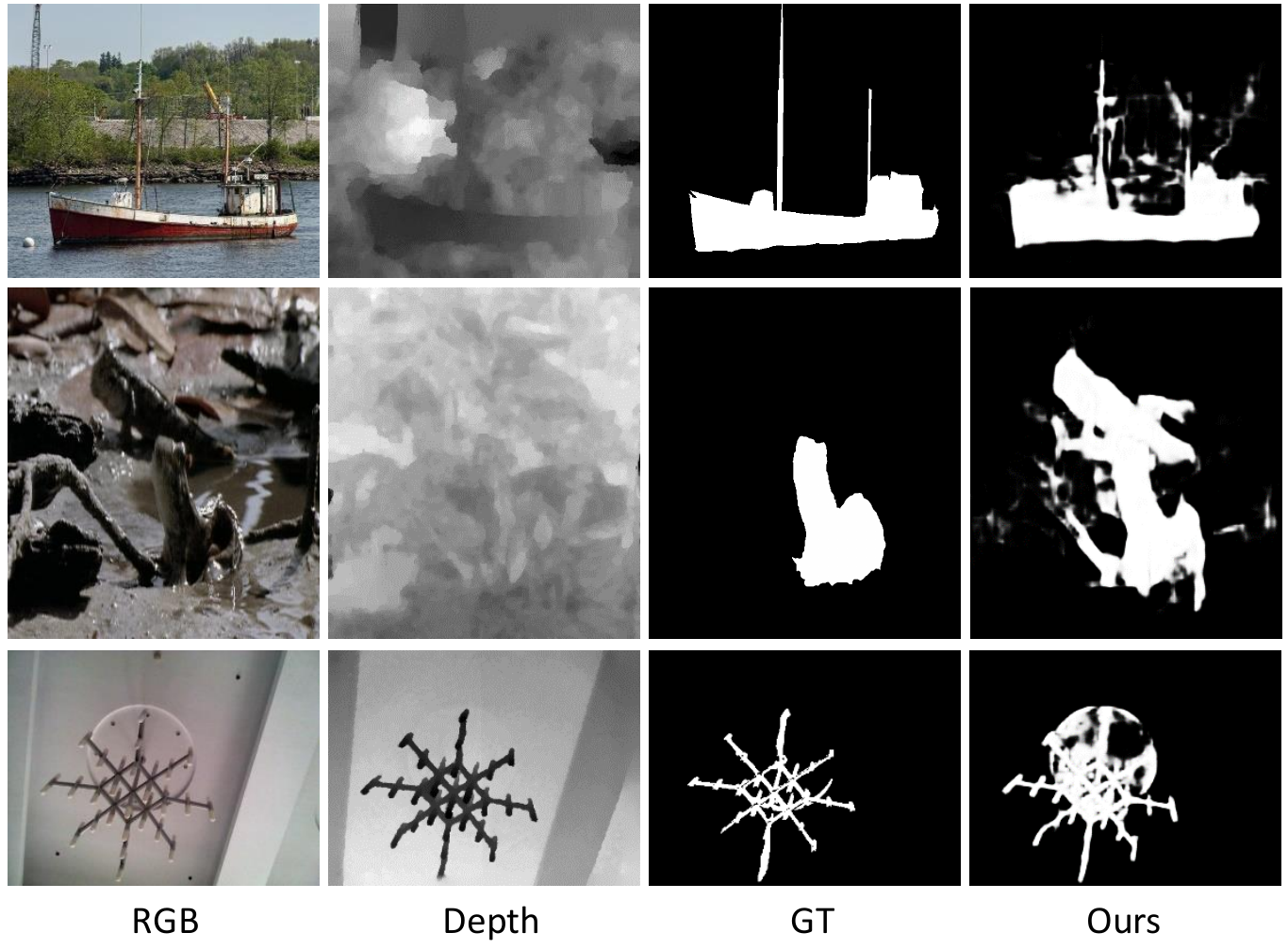}
		\caption{Some failure cases of our model.} 
		\label{fig011}
	\end{centering}
\end{figure}

4) \textbf{Object Scale}. To characterize the scale of a salient object, we compute the ratio between the size of the salient region and the whole image. Here three types of object scales can be defined: 1) when the ratio is less than 0.1, termed ``small"; 2) when the ratio is larger than 0.4, termed ``large"; and 3) when the ratio is in the range of $[0.1, 0.4]$, termed ``medium". To evaluate different methods in handling scale variation, we construct a hybrid dataset with 2,444 images collected from STERE \cite{niu2012leveraging}, NLPR \cite{peng2014rgbd}, SSD \cite{zhu2017three}, DES \cite{cheng2014depth}, and SIP \cite{fan2019rethinking}. Fig.~\ref{fig010} shows the comparison results of the attribute-based evaluation \emph{w.r.t.} scales of the salient object(s). From the results, we can see that all comparison methods obtain better performance in detecting small salient objects while they obtain relatively worse performance in detecting large salient objects. Besides, the most recent models, \ie, JL-DCF, DANet, PGAR, and our model, obtain promising performance.

\subsection{Failure Cases and Discussion}

The proposed \ours~has shown good RGB-D saliency detection performance in most cases. However, our model fails to detect salient objects when dealing with some challenging scenes such as complex background and low-quality depth. Some failure cases of our model are shown in Fig.~\ref{fig011}. In the first row, we can see that the depth quality is very poor, which makes our model can only roughly locate the boat without fine details. Thus, it is helpful to enhance or filter depth maps for boosting the saliency detection performance. In the second row, the annotated salient object has a similar appearance to other objects in the scene, thus it is challenging to accurately detect the salient object. In the third row, we can see that the object has fine details, our model only locates the main regions without fine details. Thus, there is still considerable room for improving our model to handle some scenes with fine structures in further work.

\subsection{Application for RGB-D Camouflaged Object Detection}
\label{COD_enten}

The proposed \ours~is originally designed for the RGB-D SOD task, which can be easily extended to other related RGB-D tasks, \eg, RGB-D based camouflaged object detection (COD). The aim of COD is to identify objects that are ``seamlessly” embedded in their background surroundings. Thus, it is a very challenging task due to the high intrinsic similarities between the target object and the background \cite{fan2021concealed,sun2021context,lim2021}. A recent research \cite{zhang2021depth} suggests that depth can also provide useful spatial information to boost COD performance. Thus, we extend the proposed \ours~ to the RGB-D COD task.

\begin{table}[t!]
  \centering
  \renewcommand{\arraystretch}{1.25}
  \caption{Comparison results of different camouflaged object detection models on benchmark datasets using two widely used evaluation metrics (\ie, $S_{\alpha}$ \cite{fan2017structure} and $\mathcal{M}$ \cite{perazzi2012saliency}).
  ``$\uparrow$`` \& ``$\downarrow$'' indicate that larger or smaller is better. 
  }
  \scriptsize
  \begin{tabular}{r|p{0.4cm}p{0.4cm}|p{0.5cm}p{0.5cm}|p{0.5cm}p{0.5cm}}
  \hline\toprule

    &\multicolumn{2}{c|}{CHAMELEON}
    &\multicolumn{2}{c|}{CAMO}
    &\multicolumn{2}{c}{COD10K}\\

    \textbf{Model}
    &$S_{\alpha}\uparrow$ &$\mathcal{M}\downarrow$
    &$S_{\alpha}\uparrow$ &$\mathcal{M}\downarrow$
    &$S_{\alpha}\uparrow$ &$\mathcal{M}\downarrow$ \\

    \midrule
    FPN~\cite{fpn}

    & 0.794  & 0.075
    & 0.684  & 0.131
    & 0.697  & 0.075 \\

    MaskRCNN~\cite{maskrcnn}
    & 0.643  & 0.099
    & 0.574  & 0.151
    & 0.613  & 0.080 \\

    PSPNet~\cite{pspnet}
    & 0.773 & 0.085
    & 0.663 & 0.139
    & 0.678 & 0.080 \\

    PiCANet~\cite{picanet}
    & 0.769 & 0.085
    & 0.609 & 0.156
    & 0.649 & 0.090 \\

    BASNet~\cite{basnet}
    & 0.687 & 0.118
    & 0.618 & 0.159
    & 0.634 & 0.105 \\

    PFANet~\cite{pfanet}
    & 0.679 & 0.144
    & 0.659 & 0.172
    & 0.636 & 0.128 \\

    CPD~\cite{cpd}
    & 0.853 & 0.052
    & 0.726 & 0.115
    & 0.747 & 0.059 \\

    EGNet~\cite{egnet}
    & 0.848 & 0.050
    & 0.732 & 0.104
    & 0.737 & 0.056 \\

    SINet~\cite{fan2021concealed}
    & 0.869 & 0.044
    & 0.751 & 0.100
    & 0.771 & 0.051 \\

    \midrule
    DANet~\cite{zhao2020single}
    & 0.874 & 0.043
    & 0.752 & 0.100
    & 0.765 & 0.051 \\

    HDFNet~\cite{pang2020hierarchical}
    & 0.875 & 0.032
    & 0.778 & 0.085
    & 0.779 & 0.045 \\

    \midrule

    \ours~(Ours)
    &  \textbf{0.895}  & \textbf{0.027}
    &  \textbf{0.795}  & \textbf{0.082}
    &  \textbf{0.797}  & \textbf{0.042} \\

  \bottomrule
  \hline
  \end{tabular}\label{tab7}
\end{table}

\textbf{Dataset}. We conduct this extension experiment on three public benchmark datasets for camouflaged object detection, including 1) CHAMELEON~\cite{fan2021concealed} dataset, it consists of 76 camouflaged images, 2) CAMO~\cite{anet2019} dataset, it has $1,250$ images ($1,000$ for training, $250$ for testing) with 8 categories, and 3) COD10K~\cite{fan2021concealed} dataset, it consists of $5,066$ camouflaged images ($3,040$ for training, $2,026$ for testing) with 5 super-classes and 69 sub-classes. Following the same setting in~\cite{fan2020camouflaged}, we divide the training and testing sets and then train our model using the training set.

\begin{figure*}[t]
\centering
\begin{overpic}[width=0.99\linewidth]{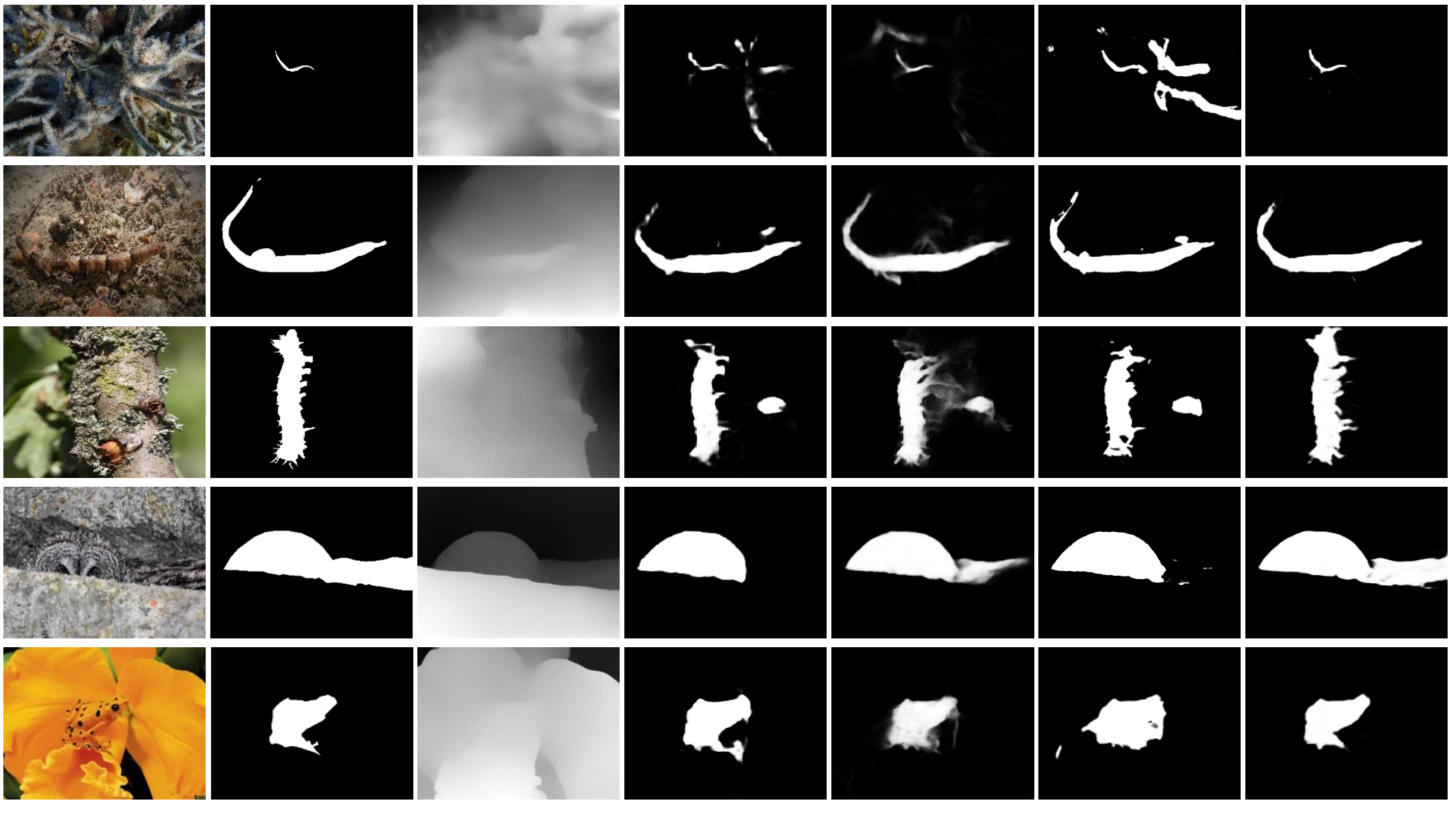}
\put(6, 1.0){\footnotesize RGB}
\put(20,1.0){\footnotesize GT}
\put(34,1.0){\footnotesize Depth}
\put(48,1.0){\footnotesize SINet}
\put(62,1.0){\footnotesize DANet}
\put(76,1.0){\footnotesize HDFNet}
\put(90,1.0){\footnotesize SPNet}
\end{overpic}

\caption{COD results of our SPNet and three state-of-the-art COD methods (\ie, SINet \cite{fan2020camouflaged}, DANet~\cite{zhao2020single}, and HDFNet~\cite{pang2020hierarchical}).}  \vspace {-0.25cm}
    \label{fig012}

\end{figure*}
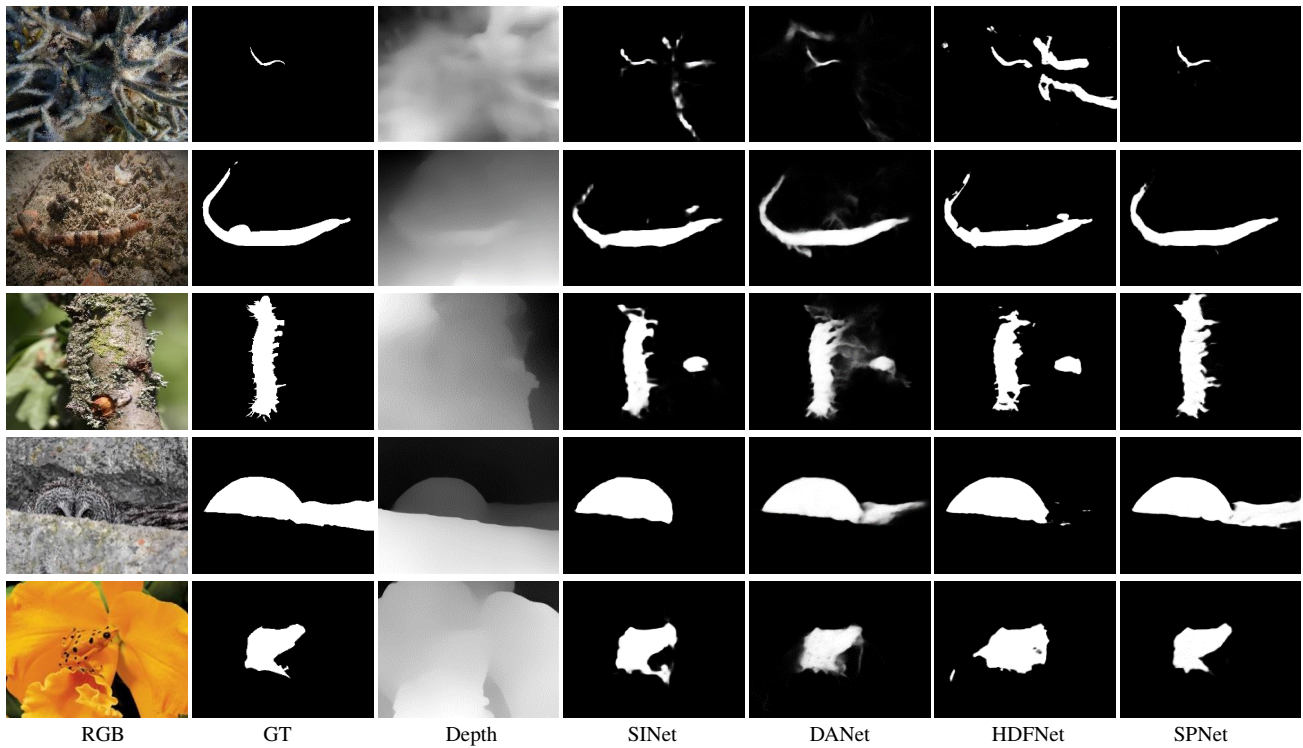

\textbf{Comparison Methods}. We compare with some existing COD models, including FPN~\cite{fpn}, MaskRCNN~\cite{maskrcnn}, PSPNet~\cite{pspnet}, PiCANet~\cite{picanet}, BASNet~\cite{basnet},  PFANet~\cite{pfanet},  CPD~\cite{cpd}, EGNet~\cite{egnet}, and SINet~\cite{fan2020camouflaged}. Note that the results for the above results are collected from the work ~\cite{fan2020camouflaged}. Since there are fewer works designed for RGB-D camouflaged object detection, we compare two recent RGB-D salient object detection methods, \ie, DANet~\cite{zhao2020single}, and HDFNet~\cite{pang2020hierarchical}, in this experiment. We re-train the two RGB-D SOD models and our model using RGB and depth images.

\textbf{Results}. Table~\ref{tab7} shows the quantitative results of different COD methods on three public datasets. From the results, it can be observed that our model performs better than other comparison COD methods. Besides, it is worth noting that our model and two RGB-D COD methods with using depth cues perform better than other methods without using them, which indicates the depth cues can provide spatial information to improve the COD performance. Fig.~\ref{fig012} shows the qualitative results of different COD methods. Compared with other COD models, we can see that our \ours~ can achieve better visual effects by detecting more accurate boundaries of camouflaged objects.

\section{Conclusion}

In this paper, we present a novel RGB-D salient object detection framework, termed \ours. Different from most existing RGB-D SOD methods, which mainly focus on learning shared representations, our \ours~ not only explores the shared cross-modal information but also compensates modality-specific characteristics to improve the SOD performance. To learn the shared representations for the two modalities, we introduce a cross-enhanced integration module (CIM) to fuse the cross-modal features, and the output of each CIM can be propagated to the next layer for exploring rich cross-level information. Besides, we adopt a multi-modal feature aggregation (MFA) module to integrate the learned modality-specific features for enhancing the complementary multi-modal information. Extensive results on benchmark datasets show the effectiveness of our model against other state-of-the-art RGB-D SOD methods. Moreover, we thoroughly validate the effectiveness of key components in our framework, and an attribute-based evaluation is conducted to study the performance of many cutting-edge RGB-D SOD approaches under different challenging factors. Finally, we extend the proposed \ours~to the recently proposed RGB-D camouflaged object detection task, and the effectiveness has also been validated. \\

{\small
\bibliographystyle{ieee_fullname}
\bibliography{egbib}
}

\end{document}